\documentclass[10pt,twocolumn,letterpaper]{article}

\usepackage[pagenumbers]{cvpr}              %

\newcommand{\name}{{ConText-CIR}}

\usepackage[normalem]{ulem}
\useunder{\uline}{\ul}{}
\usepackage{graphicx} %
\usepackage{booktabs}
\usepackage{multirow}
\usepackage{arydshln}
\usepackage{amssymb}
\usepackage{pifont}
\usepackage{pgfplots}
\pgfplotsset{compat=1.18}

\definecolor{cvprblue}{rgb}{0.21,0.49,0.74}
\usepackage[pagebackref,breaklinks,colorlinks,allcolors=cvprblue]{hyperref}
\usepackage[splitrule]{footmisc} %
\usepackage{multirow}
\usepackage[subtle]{savetrees}
\usepackage{siunitx}
\usepackage{array}
\usepackage{makecell}
\usepackage{float}
\usepackage{subcaption}
\usepackage{placeins}

\def\paperID{13770} %
\def\confName{CVPR}
\def\confYear{2025}

\title{{\name}: Learning from Concepts in Text for Composed Image Retrieval}

\author{
    \begin{tabular}{c}
        Eric Xing$^{1}$ \hspace{1em} Pranavi Kolouju$^{2}$ \hspace{1em} Robert Pless$^{3}$ \hspace{1em} Abby Stylianou$^{2}$ \hspace{1em} Nathan Jacobs$^{1}$ \\
        $^{1}$Washington University in St. Louis \\
        $^{2}$Saint Louis University \\
        $^{3}$The George Washington University \\
        \small \texttt{\{e.xing,jacobsn\}@wustl.edu, \{pranavi.kolouju,abby.stylianou\}@slu.edu, pless@gwu.edu}
    \end{tabular}
}

\begin{document}
\maketitle

\begin{abstract}

Composed image retrieval (CIR) is the task of retrieving a target image specified by a query image and a relative text that describes a semantic modification to the query image. Existing methods in CIR struggle to accurately represent the image and the text modification, resulting in subpar performance. To address this limitation, we introduce a CIR framework, {\name}, trained with a Text Concept-Consistency loss that encourages the representations of noun phrases in the text modification to better attend to the relevant parts of the query image. To support training with this loss function, we also propose a synthetic data generation pipeline that creates training data from existing CIR datasets or unlabeled images. We show that these components together enable stronger performance on CIR tasks, setting a new state-of-the-art in composed image retrieval in both the supervised and zero-shot settings on multiple benchmark datasets, including CIRR and CIRCO. Source code, model checkpoints, and our new datasets are available at \url{https://github.com/mvrl/ConText-CIR} .

\end{abstract}

\section{Introduction}
\label{sec:intro}

Traditional image retrieval is a longstanding problem in computer vision~\cite{survey} with a diverse set of applications, including visual search~\cite{shankar2017deeplearningbasedlarge}, image geolocalization~\cite{zeng2023geo}, medical imaging~\cite{SUDHISH2024105620}, etc. There are two standard approaches to the image retrieval task: image-based~\cite{zhang2024irgengenerativemodelingimage, asymmetric}, where the input is an image and the task is finding images that are visually similar, and text-based~\cite{sultan2023exploring, long2024cfirfasteffectivelongtext}, where the input is natural language, and the task is finding images that match the text. Image embedding models make the image-based approach possible, and vision-language models such as CLIP~\cite{pmlr-v139-radford21a}, which have aligned image and text representations, have facilitated the text-based approach. However, both approaches have limitations: images are typically complex and contain a wide variety of objects, making it difficult to encode specific retrieval criteria in an image alone; conversely, it is difficult to specify complex visual information in text alone, as describing details of color patterns and the shapes of interest objects is often difficult and imprecise with text.

\begin{figure}
    \centering
    \includegraphics[width=\linewidth]{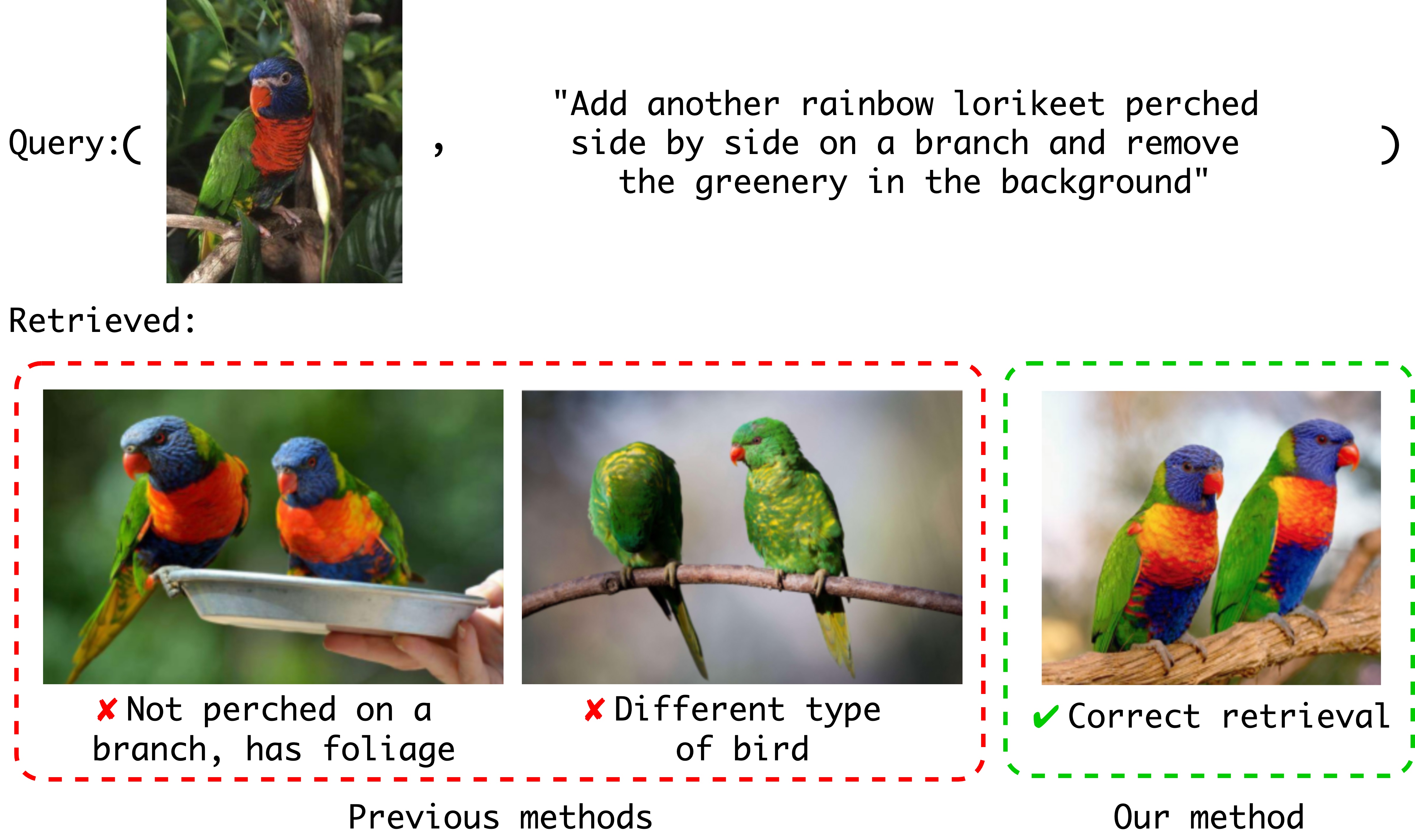}
    \caption{A failure case of current composed image retrieval methods. Previous methods do not accurately capture the two conditions specified by the text. }
    \label{fig:limitations-current}
\end{figure}

Recent work in multi-modal image retrieval aims to mitigate these inherent limitations of uni-modal queries by introducing the composed image retrieval (CIR) problem~\cite{vo2019composingempiricalodyssey,10219821_vl_alignment, Wen_2023, jiang2024dualrelationalignmentcomposed}. CIR involves retrieving a target image specified by a reference image and a text modification to the reference image. Incorporating open-ended text as an additional search criterion allows the inclusion of diverse natural language concepts, modifiers, and search specifications with images to produce a flexible retrieval formulation. Current work in CIR, however, fails to model the interaction of longer, richer texts with images. Existing approaches and datasets focus on simple text modifiers, with a straightforward modification, most often to foreground objects. Real-world queries, however, are often complex, multi-attribute, and may express modifications to background objects. Figure~\ref{fig:limitations-current} demonstrates the limitations of current CIR methods, where the query includes an image of a ``rainbow lorikeet'' and text saying to ``add another lorikeet on the branch and to remove the greenery in the background". Current methods tend to retrieve results that are correlated with the concepts specified in the text modifier (adding a second lorikeet, but not changing the background, or adding a different type of bird), but do not completely capture the entire relationship between image query and text modification.

We introduce a novel CIR framework, {\name}: Learning from \textbf{Con}cepts in \textbf{Text} for \textbf{C}omposed \textbf{I}mage \textbf{R}etrieval, that addresses the limitations of previous methods in handling complex, multi-attribute texts. Existing vision-language models like CLIP demonstrate reasonable alignment between text and image concepts when the text is simple. For multi-attribute texts, however, the correspondence between the text and relevant image features often becomes inconsistent. Our proposed concept-consistency (CC) loss addresses this issue by enforcing consistent cross-attention between specific noun phrases and corresponding image regions. Specifically, our loss function encourages the attention weight of a noun phrase in the context of a full sentence to match its weight when evaluated independently, thus reducing contextual interference in long text phrases.

Our framework is trained using a standard CIR formulation and may be easily extended to new datasets as methods and efforts for dataset generation develop. We aim to address the current limited performance of state-of-the-art CIR methods on queries with multi-attribute modifier text, increasing the applicability of CIR models on complex retrieval tasks and setting a new state-of-the-art on current benchmarks.  Our specific contributions are the following:
\begin{itemize}
    \item We introduce a new CIR framework, {\name}, demonstrating state-of-the-art composed image retrieval performance on the CIRR and CIRCO benchmarks.
    \item We qualitatively show that our concept-consistency loss helps the underlying encoders to learn more specific object-centric representations. 
    \item We introduce a synthetic data generation pipeline to generate multi-attribute text annotations for existing small CIR datasets or to generate new CIR datasets for new domains.
    
\end{itemize}

\begin{figure*}[!t]
    \centering
    \includegraphics[width=\textwidth]{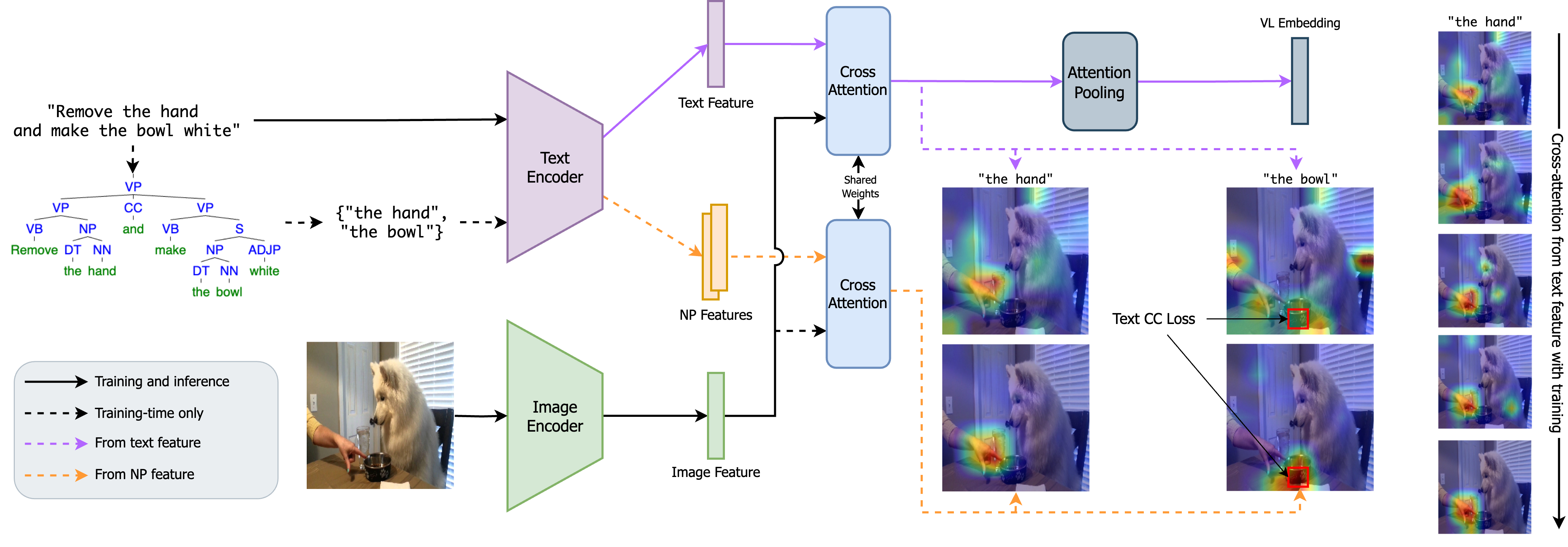}
    \caption{The overall architecture of our approach, \name. The framework guides attention to the related image regions by penalizing large differences between attention maps resulting from concept-specific and whole-text representations for each noun phrase. During inference, our method operates efficiently using a simple cross-attention mechanism to combine image and text features. The right side of the figure shows that the cross-attention between the concept ``the hand" \textit{from the representation of the entire text} and the image query converges to the local region around the hand with very little spurious attention. }
    
    \label{fig:arch}
\end{figure*}

\section{Background \& Related Work}
Our work sets a new state-of-the-art for composed image retrieval. In this section, we summarize the most relevant background and closely related work.

\subsection{Composed Image Retrieval}

Modern methods in composed image retrieval primarily use multimodal fusion methods to combine representations of the query image and relative text. This produced vision-language embedding may then be used to perform a database lookup to retrieve likely candidate images~\cite{vo2019composingempiricalodyssey, delmas2022artemis, lfclip_lfblip_9879245, sun2024trainingfreezeroshotcomposedimage_grb_tfcir, Liu_2021_ICCV_cirplant, baldrati2023composed_clip4cir}. The success of pre-trained vision language models~\cite{radford2021learning, li2022blip} has motivated a number of methods that utilize these performant models to extract powerful features to describe the query image and the modification text. Work in CIR has explored a variety of methods, including sophisticated attention-based mechanisms~\cite{9157634_vislingattnlearningchen, zhang2024magiclens}, denoising methods~\cite{gu2024compodiff}, and simple interpolation-based approaches~\cite{jang2024sphericallinearinterpolationtextanchoring_slerp}, to fuse these representations.

Powerful generative VLMs~\cite{liu2023visualinstructiontuning_llava, dai2023instructblipgeneralpurposevisionlanguagemodels} have also inspired training-free methods to perform CIR and composed video retrieval~\cite{ventura24covr, 10685001_covr-2_blip} has been used as a proxy task to obtain powerful visuolingual understanding from large video datasets~\cite{Bain21_webvid}. In another vein, textual inversion-based methods learn pseudoword representations of the image query~\cite{baldrati2023zeroshot_searle, saito2023pic2word} in the text space of a pre-trained language model. These methods also benefit from learning greater image concept-text alignment by decomposing concepts in an image into multiple pseudo-words~\cite{gu2024lincir}. Other methods~\cite{10568424_align_retrieve, Wan_2024_CVPR} explicitly aim to learn multimodal alignments but do not learn the grounding of fine-grained text concepts with their corresponding image features. Some methods~\cite{song2024syncmask, 10219821_vl_alignment} aim to learn finer-grained alignments between text and image by learning cross-modal masks or differences between features for fashion-centric retrieval. However, all of these methods suffer from poor concept-level alignment between precise text concepts and their corresponding image features due to a lack of structured conceptual guidance, inheriting only coarse-level alignment from contrastive pretraining~\cite{DBLP:journals/corr/abs-2111-07783_filip, inproceedings_fdtpaper, song2023FDalign}.

\subsubsection{Image-Text Datasets} 
Early work in this area focused on visual question answering regarding image composition. Data sets like NLVR~\cite{suhr2017corpus}, CLEVR~\cite{johnson2017clevr}, and GQA~\cite{hudson2019gqa} focused on simple synthetic and real images with questions and ground-truth answers about shape, size, and object configurations. The simplicity of these datasets led to the development of NLVR2~\cite{suhr2018corpus}, a larger dataset of natural image pairs with more complex compositional queries. Fashion-IQ~\cite{guo2019fashion}, a fine-grained dataset for fashion-focused composed image retrieval, was one of the first true CIR datasets, providing image pairs with ``relative'' captions describing specific modifications to clothing, along with a baseline multimodal transformer approach for retrieval.

The CIRR dataset~\cite{Liu_2021_ICCV_cirplant}, derived from NLVR2, includes human annotations focused on modifications to target images, facilitating CIR tasks. Although the most commonly evaluated dataset, the CIRR dataset has some limitations that have been pointed out by others~\cite{baldrati2023zeroshot_searle}, including being limited to image pairs from NLVR2, having many captions that do not relate to the query at all, and only including annotations that describe a single modification to a foreground object. CIRCO~\cite{baldrati2023composed_clip4cir} is a test-only dataset that consists of human-generated annotations for image pairs mined from the MS-COCO dataset~\cite{lin2015microsoftcococommonobjects} and notably contains more complex text annotations and multiple targets per annotation. While these annotations are high quality and more complex than the CIRR annotations, it does not have a training dataset and can only be used to evaluate the performance of models trained on other data. More recent datasets like LaSCo~\cite{levy2024data} produce larger training datasets of synthetic annotations by mining examples from larger existing annotated datasets like VQA2.0~\cite{balanced_vqa_v2}. The SynthTriplets18M uses InstructPix2Pix~\cite{brooks2022instructpix2pix} to generate images based on automatically generated text prompts~\cite{gu2024compodiff}. There are also domain-specific CIR datasets, such as the Birds-to-Words dataset~\cite{forbes2019neural} and the PatterCom remote sensing dataset~\cite{psomas2024composed}, and video retrieval datasets~\cite{ventura24covr, ventura24covr2}. In general, existing CIR datasets are limited by the quality of their text annotations, motivating our work to leverage strong multimodal language models with refined prompts for generating high-quality CIR data.

\subsection{Visual Grounding}
Our proposed {\name} framework enhances visual grounding, or the association between image regions and specific textual concepts, improving query-result alignment. Recent advancements in visual grounding span areas such as visual question answering (VQA), text-guided image editing, and text-guided segmentation. In VQA, Contrastive Region Guidance (CRG) strengthens visual grounding by comparing model outputs with and without text prompts to focus on relevant image areas without extra training~\cite{CRG}. In text-guided image editing, work has focused on enhancing localization via cross-attention mechanisms \cite{NEURIPS2023_5321b1da, simsar2023limelocalizedimageediting, hertz2022prompt}, addressing weak grounding due to loose text-to-attention maps. For Prompt2Prompt\cite{hertz2022prompt}, cross-attention is refined to control pixel influence per token. LIME employs cross-attention regularization, penalizing irrelevant scores within the ROI, highlighting relevant tokens while downplaying unrelated ones~\cite{simsar2023limelocalizedimageediting}. In image segmentation, models like DINO~\cite{caron2021emerging} and SegmentAnything~\cite{kirillov2023segany} show impressive results but lack \textit{semantic} segmentation capabilities, leading to grounding efforts with natural language, such as Grounding DINO~\cite{liu2023grounding} and Grounded-SAM~\cite{ren2024grounded}. Other approaches like LSeg~\cite{li2022languagedriven} train image encoders to align local embeddings to semantic text embeddings, while CMPC~\cite{huang2020referring} enhances grounding by utilizing entity relationships in textual prompts.

\section{Methodology}
\label{sec:methodology}

We propose {\name} for training a composed image retrieval model, including a Text Concept-Consistency loss that improves the visual grounding of text concepts in queries. %

\subsection{Problem Setting}
Composed image retrieval is formulated as finding a target image in a database of gallery images that best matches a query specified by a query image and relative text. In practice, this involves pre-indexing the database of gallery images with some embedding function $f$ and retrieving candidates using a representation of the image-text query mapped into the same space.

Formally, we aim to learn a mapping from a query image, relative text tuple $(I_q, T)$ to a target image $I_T$.  We aim to learn a multimodal representation network $r=f(I_q, T)$ that produces a vector to query a database of images. %

\subsection{The {\name} model}

Our multimodal model consists of pair of encoders for image and text, $\phi_I$, $\phi_T$, a cross attention mechanism $\phi_A$, and an attention pooling mechanism $\phi_P$. Figure ~\ref{fig:arch} shows a diagram of our pipeline. The final representation of a query pair consisting of an image and modification text is created using cross-attention to fuse the representations of text and image,
$$CrossAttn(I, T)= \phi_A(\phi_I(I), \phi_T(T)),$$
followed by attention pooling to obtain the final vision-language feature $r$:
$$r=f(I, T) = \phi_P(CrossAttn(I, T)).$$
This model is trained with two loss functions: a contrastive loss between query image-language representations and target image representations and a novel loss that explicitly encourages the cross-attention between text concepts to attend to their relevant image tokens. We describe each in the following sections.

\subsection{Text Concept-Consistency Loss}

The novel Text Concept-Consistency (Text CC) loss function encourages stronger consistency between noun phrases in the text and the correct image region.  Visual language models have a strong capacity to find relationships between simple text and image regions~\cite{li2021groundedglip,liu2023grounding}, however, representations extracted from multi-concept texts by language models may produce representations that have poorer alignment between particular concepts and their relevant image regions, as global contrastive pretraining methods do not ensure fine-grained alignment. Due to the complex nature of text-image interaction in CIR (involving changing image attributes, removing subjects, etc.), we aim to learn strong concept-level alignments between text and image representations. Following work in linguistics~\cite{MURPHY1990259_nounphrase} we use noun phrases (NP) to represent concepts in a text.

Before training, we extract at most $l$ noun phrase (NP) constituents in each text $T$. For each noun phrase ${NP}_i$ we consider the in-context cross-attention of the tokens:
$$CrossAttn(I,T_{NP_i})$$
between the tokens for the $i$-th noun-phrase in the context of the original text $T$, and the isolated cross-attention:
$$CrossAttn(I,NP_i)$$
between the tokens for the $i$-th noun phrase when represented \textit{by itself} with the text encoder.  With these components, we define a loss function to encourage the in-context cross-attention map to match the isolated cross-attention map generated from the concept-specific embedding, summed over all noun phrases extracted from the sentence:
\begin{equation*}
\begin{split}
\mathcal{L}_{cc} = \sum_{i = 1}^{l} 
\text{ReLU}\big( & \; CrossAttn(I,T_{NP_i}) \\
& - CrossAttn(I,NP_i) - \epsilon \big)
\end{split}
\end{equation*}
A slack variable $\epsilon$ permits some difference in the magnitude of cross-attention values between in-context and isolated noun-phrase embeddings and is set as a hyperparameter. Intuitively, learning from cross-attentions produced from concept-only embeddings will improve the grounding of concepts to images as the concept-specific embeddings cannot be confused with attributes from other concepts or suppressed by other concepts in the text, leading to focused grounding.

We also enforce the contrastive loss between representations of the query with text modification $r^q=f(I_q, T)$ and the target image $r_n^t=f(I_t, ``\ \ ")$ where $I_q,I_t$ are the query and target images respectively. The target image is encoded with the empty text so that the query vision-language embedding and target image embedding lie in the same space. Following MagicLens~\cite{zhang2024magiclens}, we also include the representation of the query image as an additional negative example, $r_n^{q-}=f(I_q, ``\ \ ")$ encoding the query image with the empty text. So, the contrastive loss is defined as follows for the $n$-th $(I_q, T, I_t)$ training triplet:
$$\mathcal{L}_{cont} = -\log\frac{e^{\text{sim}(r^q_n, r^t_n) / \tau}}{e^{\text{sim}(r^q_n, r^{q-}_n) / \tau}+\sum_{i=1}^N e^{\text{sim}(r^q_n, r^t_i) / \tau}}$$
where $\text{sim}(r_a,r_b)$ denotes the cosine similarity $\frac{r_a\cdot r_b}{||r_a||||r_b||}$, $N$ is the training batch size, and $\tau$ is a temperature hyperparameter to scale the distribution of logits.  The total loss is given as follows, where $\lambda$ scales the weight given to each loss:
$$\mathcal{L}_{tot} = \mathcal{L}_{cont} + \lambda \mathcal{L}_{cc}.$$

\subsection{{\name} Inference}
To perform image retrieval with respect to a gallery of candidate images, each candidate image $C_i$ is encoded with a blank text vector as: $r_i = f(C_i, ``\ \ ")$.  Given a query text pair $I_q, T$, each candidate image $C_i$ is scored based on the similarity between $r_i$ and $f(I_q, T)$; we report results in this paper on top-$k$ results for various values of $k$.

\subsection{Automated Data Generation Pipeline}

\begin{figure}[t]
    \centering
    {\scriptsize %
    \setlength{\tabcolsep}{1pt} %
    \renewcommand{\arraystretch}{3} %
    \begin{tabular}{cc>{\centering\arraybackslash}m{2cm}>{\centering\arraybackslash}m{3.2cm}} %
    \textbf{Query Image} & \textbf{Target Image} & \textbf{Text Difference} & \textbf{Issue} \\ \toprule
        \raisebox{-.25in}{\includegraphics[height=.5in]{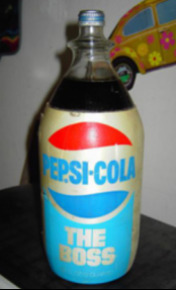}} & \raisebox{-.25in}{\includegraphics[height=.5in]{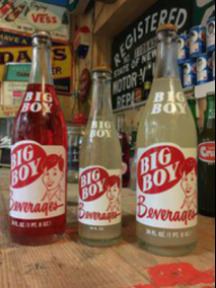}} & ``show three bottles of soft drink''~\cite{Liu_2021_ICCV_cirplant} & Query photo is unnecessary \\
        \raisebox{-.25in}{\includegraphics[height=.5in]{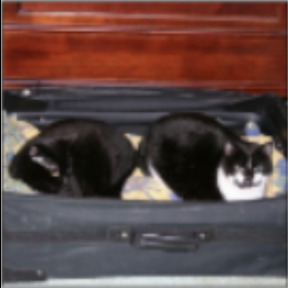}} & \raisebox{-.25in}{\includegraphics[height=.5in]{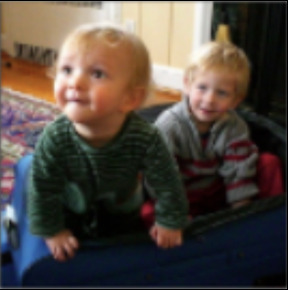}} & ``has two children instead of cats''~\cite{baldrati2023zeroshot_searle} & Images are not visually similar \\
        \raisebox{-.25in}{\includegraphics[height=.5in]{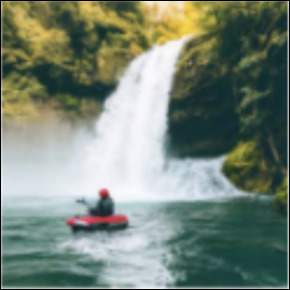}} & \raisebox{-.25in}{\includegraphics[height=.5in]{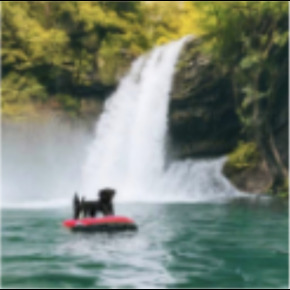}} & ``Have the person be a dog''~\cite{gu2024compodiff} & Images are too visually similar \\
        \raisebox{-.25in}{\includegraphics[height=.5in]{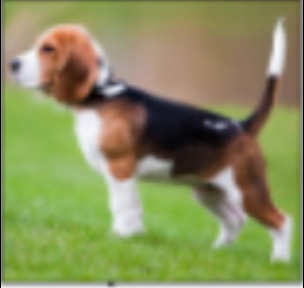}} & \raisebox{-.25in}{\includegraphics[height=.5in]{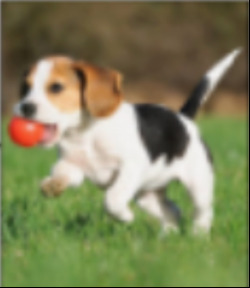}} & ``Add a red ball''~\cite{baldrati2023composed_clip4cir} & Modification is very simple
    \end{tabular}
    }
    \caption{\label{fig:data_issues} Qualitative issues with existing CIR datasets.}
\end{figure}

CIR datasets can be divided into two categories: manually generated and automatically generated. Manually generated datasets like CIRR~\cite{Liu_2021_ICCV_cirplant} ask human annotators to describe the differences between image pairs (though the pairs are often extracted from existing datasets). Automatically generated datasets tend to leverage existing labeled data to mine CIR queries (e.g., LaSCo~\cite{levy2024data}), or use image generation tools to produce images that match specific modification text, as in the case of the SynthTriplets18M dataset~\cite{gu2024compodiff}. There are a variety of problems with these existing datasets regardless of the method of generation, including queries where the text on its own is sufficient to find the target image and issues with the degree of image similarity in the queries. Across existing datasets, the modifications are also typically simple, focusing on a single change to a foreground object. We show examples of these issues in Figure~\ref{fig:data_issues}.

To address these issues, and facilitate the generation of new CIR datasets with realistically complicated text modifications, we introduce a novel pipeline good4cir~\cite{kolouju2025good4cir} leveraging a large language model (specifically OpenAI's GPT-4o) to produce CIR triplets.\footnote{The MagicLens paper presents a similar idea of using the PaLM2 LLM to create a CIR dataset, and reports results on a generated training dataset consisting of 36.7 million CIR triplets~\cite{zhang2024magiclens}. As of November 2024, this dataset is unavailable and no code has been shared to replicate it~\cite{magiclens_github}.} In this pipeline, we assume that there is an existing list of related images--this can come from either an existing CIR dataset with low quality annotations, or from a new domain with image pairs. We then decompose the CIR triplet generation task into four smaller, targeted tasks to mitigate hallucination and encourage the generation of fine-grained descriptors~\cite{hallucination}.

First task, we generate a list of objects found in the query image, and have the LLM describe each object with a list of descriptors. Next, we pass this list of objects and associated descriptors to the LLM along with the target image, and ask the LLM to generate a list of objects and descriptors for the target image according to the following criteria:

\begin{itemize}
    \item If a new object is introduced in the target image, generate a set of descriptors for the given object. 
    \item If there is an object in the target image that matches the descriptors of an object from the query image, adopt the exact set of descriptors to ensure consistency.
    \item If there is a similar object in the query and target images, but the descriptors provided in the list for the query image do not match the appearance of the object in target image, generate a new set of descriptors for the given object. 
\end{itemize}

\begin{figure}[t]
    \centering
    {\scriptsize %
    \setlength{\tabcolsep}{1pt} %
    \renewcommand{\arraystretch}{2} %
    \begin{tabular}{cc>{\arraybackslash}m{4cm}} %
    \textbf{Query Image} & \textbf{Target Image} & \textbf{Text Differences} \\ \toprule
    \raisebox{-.4in}{\includegraphics[width=.8in]{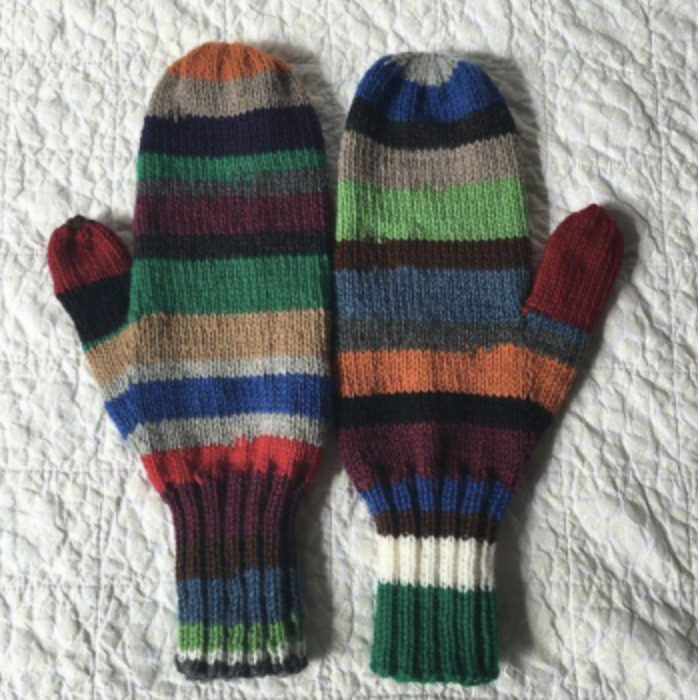}} &
    \raisebox{-.4in}{\includegraphics[width=.8in]{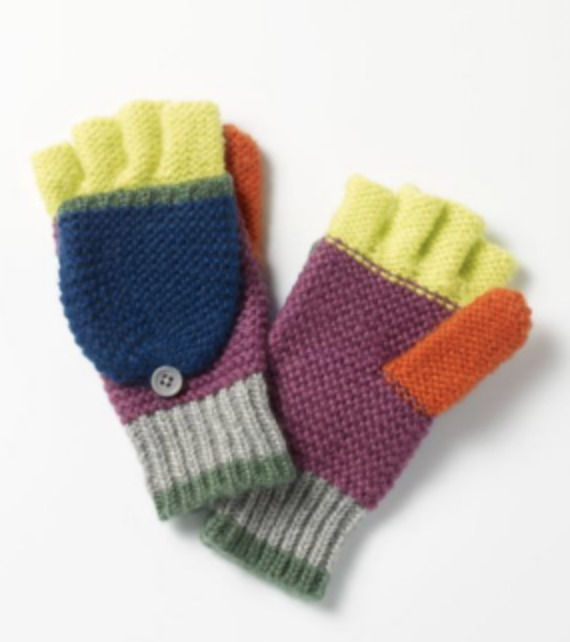}} & 
    \textbf{CIRR:}~different color pattern \newline \textbf{Ours:}~Convert the pair of mittens with multicolored stripes into fingerless gloves featuring a bright multicolored design and a convertible mitten flap with a button detail \\ 
    \raisebox{-.2in}{\includegraphics[width=.8in]{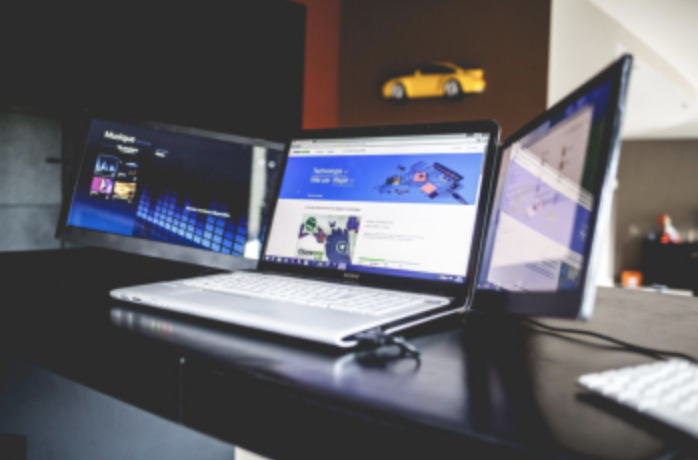}} &
    \raisebox{-.2in}{\includegraphics[width=.73in]{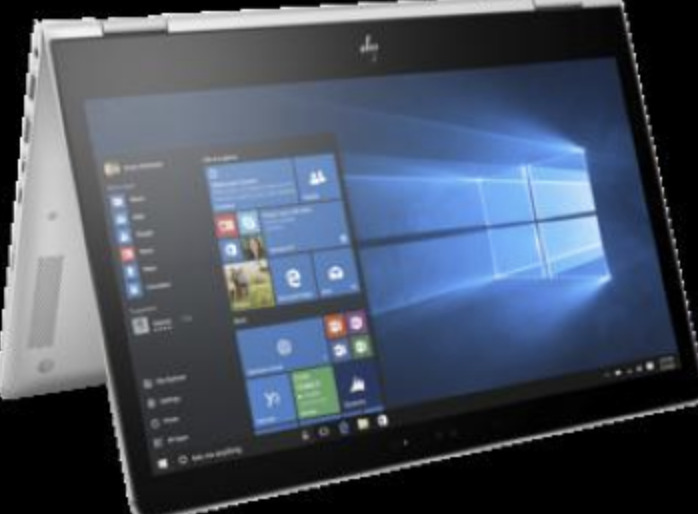}} & 
    \textbf{CIRR:}~Prop open the notebook \newline \textbf{Ours:}~Remove the left and right monitors, and replace the laptop with a 2-in-1 convertible featuring a visible hinge and silver frame \\ 
    \raisebox{-.2in}{\includegraphics[width=.73in]{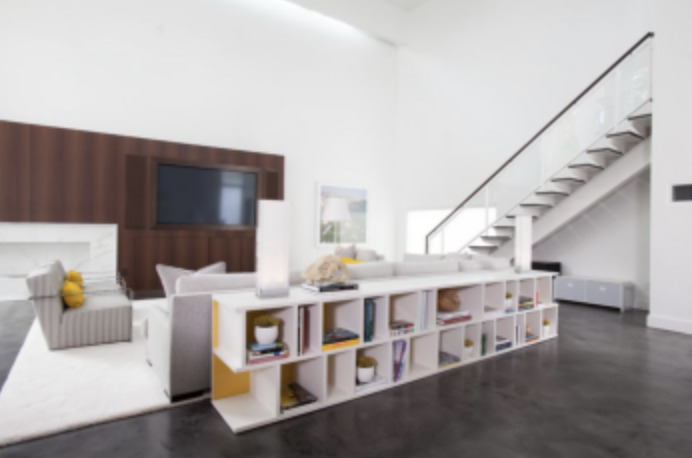}} &
    \raisebox{-.2in}{\includegraphics[width=.8in]{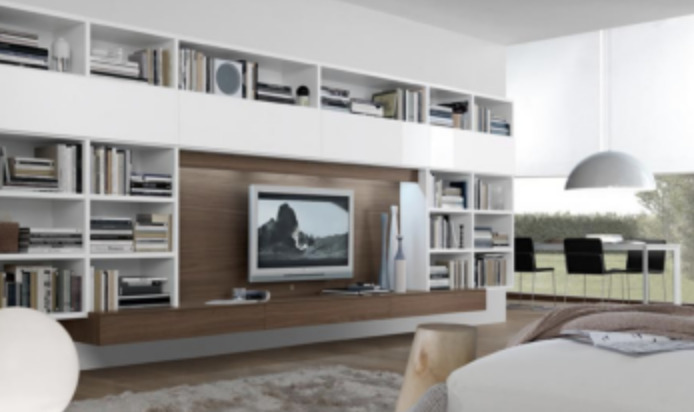}} & 
    \textbf{CIRR:}~Bigger structure and it is fixed in the wall \newline \textbf{Ours:}~Modify the bookshelf setup by integrating it with the TV unit and filling it exclusively with books \\ 
    
    \end{tabular}
    }
    \caption{Examples of original captions and rewritten CIRR captions.}
  \label{fig:cirr-and-rewrite-example}
\end{figure}
\noindent Next, we provide the LLM with both lists and have it generate a set of difference captions, each describing a single removal, addition, or modification of an object from the query image to the target image. For a given pair of images, it is possible to have a large number of modifications, each of which can be used to construct a CIR (query image, text modifier, target image) triplet. 

The final step of our pipeline is to create additional triplets by concatenating multiple text modifiers into increasingly complex queries. We show examples of generated triplets in Figure~\ref{fig:cirr-and-rewrite-example} and provide additional details and examples in the Appendix.%

\section{Experimental Setup}
\label{sec:results}

{\name} achieves significant performance improvement relative to the state-of-the-art across a broad range of standard evaluation protocols, all without increasing inference-time complexity or using a greater magnitude of training data.

\begin{table*}[t]
\centering
{\small
\begin{tabular}{@{}rll|cccc|ccc@{}}
\toprule
 &  &  & \multicolumn{4}{c|}{Recall@K} & \multicolumn{3}{c}{$\text{Recall}_{subset}\text{@K}$} \\ \midrule
Backbone & Method & Additional Data & K=1 & K=5 & K=10 & K=50 & K=1 & $K=2$ & $K=3$ \\ \midrule
\multirow{9}{*}{$\leq$ ViT-B} 
 & ARTEMIS$^*$~\cite{delmas2022artemis} & FashionIQ & 16.96 & 46.10 & 61.31 & 87.73 & 39.99 & 62.20 & 75.67 \\
& CIRPLANT~\cite{Liu_2021_ICCV_cirplant} & FashionIQ & 19.55 & 52.55 & 68.39 & 92.38 & 39.20 & 63.03 & 79.49 \\
 & LF-BLIP~\cite{lfclip_lfblip_9879245} &FashionIQ  & 20.89 & 48.07 & 61.16 & 83.71 & 50.22 & 73.16 & 86.82 \\
 & Combiner~\cite{lfclip_lfblip_9879245}$^*$ & FashionIQ & 33.59 & 65.35 & 77.35 & 95.21 & 62.39 & 81.81 & 92.02 \\
 & CLIP4CIR~\cite{baldrati2023composed_clip4cir}$^*$ & FashionIQ & 44.82 & 77.04 & 86.65 & 97.90 & 73.16 & 88.84 & 95.59 \\
 & CASE~\cite{levy2024data} & LaSCo & 48.68 & 79.98 & 88.51 & 97.49 & 76.39 & 90.12 & 95.86 \\
 & CASE~\cite{levy2024data} & LaSCo+COCO & 49.35 & 80.02 & 88.75 & 97.47 & 76.48 & 90.37 & 95.71 \\
 & {\name} (ours) & Aggregated & \textbf{49.83} & \textbf{81.54} & \textbf{89.76} & \textbf{98.95} & \textbf{76.64} & \textbf{90.69} & \textbf{96.31} \\ \midrule
\multirow{3}{*}{ViT-L} & CompoDiff~\cite{gu2024compodiff} & ST18M+LAION2B & 22.35 & 54.36 & 73.41 & 91.77 & 62.55 & 81.44 & 90.21 \\
 & CoVR-BLIP~\cite{ventura24covr} & WebVid-CoVR & 49.69 & 78.60 & 86.77 & 94.31 & 75.01 & 88.12 & 93.16 \\
 & {\name} (ours) & Aggregated & \textbf{52.65} & \textbf{83.27} & \textbf{89.51} & \textbf{98.87} & \textbf{80.32} & \textbf{92.13} & \textbf{96.08} \\ \midrule
 ViT-H  & {\name} (ours) & Aggregated & \textbf{55.24} & \textbf{84.85} & \textbf{90.75} & \textbf{98.82} & \textbf{82.96} & \textbf{93.12} & \textbf{97.04} \\ \cdashline{1-10} \addlinespace[2pt]
\multirow{3}{*}{ViT-G} & CompoDiff~\cite{gu2024compodiff} & ST18M+LAION2B & 32.39 & 57.61 & 77.25 & 94.61 & 67.88 & 85.29 & 94.07 \\
 & CoVR-2~\cite{10685001_covr-2_blip} & CC-CoIR & 50.63 & 81.04 & 89.35 & 98.15 & 76.53 & 90.43 & 96.00 \\
 & CoVR-2~\cite{10685001_covr-2_blip} & WV-CC-CoVIR & 50.43 & 81.08 & 88.89 & 98.05 & 76.75 & 90.34 & 95.78 \\
 \bottomrule
\end{tabular}
}
\caption{Composed image retrieval results on the CIRR test set. Higher is better for all metrics, best results are shown in bold. The asterisks ($^*$) in the $\leq$ ViT-B section denote that a ResNet-based backbone was used.}
\label{tab:main_cirr}
\end{table*}

\subsection{Training Data}
We train on a variety of datasets, including CIRR~\cite{Liu_2021_ICCV_cirplant} and LaSCo~\cite{levy2024data}, as well as two datasets generated using our pipeline. The first generated dataset is a rewritten version of the CIRR dataset. CIRR is one of the most commonly used CIR datasets for both training and evaluation, however, as demonstrated in Figures~\ref{fig:data_issues}~and~\ref{fig:cirr-and-rewrite-example}, the existing CIRR dataset contains relatively short and minimal captions that do not translate to domains where more detailed text queries are advantageous. Thus, we generate a rewritten CIRR dataset, CIRR$_R$, that applies the synthetic data pipeline to generate longer and more linguistically rich captions for the existing CIRR image pairs.

We additionally generate a new dataset, Hotel-CIR. This dataset is sourced from visually similar image pairs in the TraffickCam~\cite{hotels50k} hotel image database. This is a compelling domain, as the scenes are dense, with a large number of objects, and high degrees of visual similarity between images that don't necessarily come from the same class. The Appendix includes additional information regarding the CIRR$_R$ and Hotel-CIR datasets, including the exact prompts used to generate them and dataset statistics. We share these datasets to motivate further work in building CIR methods that are capable of supporting real-world queries. %

We perform ablations on the inclusion of different datasets in the training process, and ultimately train a CIR model on a combined dataset of CIRR, CIRR$_R$, LaSCo, and Hotel-CIR, which we refer to as the Aggregated dataset.

\subsection{Baseline Methods}
We evaluate our method against state-of-the-art CIR methods that pre-index a database of gallery images with vector embeddings. For fairness, we compare to models initialized with CLIP ViT-B/L or OpenCLIP ViT-H when available. We evaluate against a number of textual inversion-based methods~\cite{baldrati2023zeroshot_searle, gu2024lincir, saito2023pic2word}, late multi-modal fusion-based methods~\cite{zhang2024magiclens, jang2024sphericallinearinterpolationtextanchoring_slerp, delmas2022artemis, Liu_2021_ICCV_cirplant, lfclip_lfblip_9879245, baldrati2023composed_clip4cir}, early fusion-based methods~\cite{levy2024data}, composed video retrieval-based methods that can perform CIR~\cite{ventura24covr, 10685001_covr-2_blip}, and training-free methods~\cite{yang2024ldre, karthik2024visionbylanguage_cirevl, sun2024trainingfreezeroshotcomposedimage_grb_tfcir}. 

We evaluate our framework on the CIRR~\cite{Liu_2021_ICCV_cirplant} and CIRCO~\cite{baldrati2023zeroshot_searle} test sets. CIRR also defines a subset retrieval metric, $\text{Recall}_{subset}@K$, where the model performs retrieval among a focused small subset of images for each query. The CIRCO~\cite{baldrati2023zeroshot_searle} dataset addresses certain limitations of CIRR by increasing the gallery index size and by providing multiple ground-truth targets per image-text query. CIRCO sources images from the COCO 2017 unlabeled image set~\cite{lin2015microsoftcococommonobjects}. As there are multiple ground-truths per image-text query, we report mean average precision (mAP@K). 

Additionally, we evaluate the performance of models trained on the Aggregated dataset on the CIRCO test set and models trained with LaSCo and Hotel-CIR on the CIRR test set, a CIR task setting defined in the literature as \textit{zero-shot} CIR.  Additional zero-shot results on FashionIQ ~\cite{guo2019fashion} and ImageNet-R~\cite{hendrycks2021manyfacesofrobustness} are provided in the supplemental materials.

\subsection{Implementation Details}

We use Stanza~\cite{qi2020stanza} to perform constituency parsing over each modification text, extracting at most $l=10$ noun phrases from each parse tree. The noun phrases are extracted with a breadth-first search, so the branch-level noun phrases are prioritized to be kept over the leaf-level noun phrases.  We utilize the official image and text encoders from CLIP ViT-B and ViT-L~\cite{pmlr-v139-radford21a} and OpenCLIP ViT-H~\cite{ilharco_gabriel_2021_5143773_openclip}. We follow the attention pooling formulation of Set Transformer~\cite{lee2019set_attnpooling}. We use the AdamW optimizer~\cite{loshchilov2019decoupledweightdecayregularization} with a cosine annealed learning rate cycling between \texttt{2e-5} and \texttt{2e-7} and a weight decay of \texttt{1e-2}.

\subsection{Main Results}

Table~\ref{tab:main_cirr} gives both $\text{Recall}@K$ and $\text{Recall}_{subset}@K$ metrics on the CIRR test set. When isolating each backbone, we observe that our largest model demonstrates the highest retrieval performance across both $\text{Recall}@K$ and $\text{Recall}_{subset}@K$ metrics. Within each category of encoder size, {\name} outperforms baseline methods across all metrics, especially for models using ViT-L and $\geq$ViT-H methods. Notably, {\name} improves R@1 from 50.43 to 55.24 and R@1$_{subset}$ from 76.75 to 82.96. Unlike the second-best method CoVR-2~\cite{10685001_covr-2_blip}, {\name} does not use ViT-G, a backbone with 60\% more parameters than ViT-H, and does not pretrain on a dataset of 4.9 million samples. For the ViT-L and $\leq$ViT-B encoder sizes we also outperform baseline methods on all benchmarks.

\subsection{Zero-shot Composed Image Retrieval}
\begin{table*}[t!]
\centering
\footnotesize
\begin{tabular}{@{}rll|cccc|ccc@{}}
\toprule
 &  &  & \multicolumn{4}{c|}{Recall@K} & \multicolumn{3}{c}{$\text{Recall}_{subset}\text{@K}$} \\ \midrule
Backbone & Method & Training Data & K=1 & K=5 & K=10 & K=50 & K=1 & $K=2$ & $K=3$ \\ \midrule
\multirow{8}{*}{ViT-B} & SEARLE-OTI~\cite{baldrati2023zeroshot_searle} & ImageNet1K & 24.27 & 53.25 & 66.10 & 88.84 & 54.10 & 75.81 & 87.33 \\
 & SEARLE~\cite{baldrati2023zeroshot_searle} & ImageNet1K & 24.00 & 53.42 & 66.82 & 89.78 & 54.89 & 76.60 & 88.19 \\
 & LDRE~\cite{yang2024ldre} & - & 25.69 & 55.13 & 69.04 & 89.90 & 60.53 & 80.65 & 90.70 \\
 & CIReVL~\cite{karthik2024visionbylanguage_cirevl} & - & 23.94 & 52.51 & 66.00 & 86.95 & 60.17 & 80.05 & 90.19 \\
 & MagicLens~\cite{zhang2024magiclens} & web-scraped (36.7M) & 27.0\phantom{0} & 58.0\phantom{0} & 70.9\phantom{0} & 91.1\phantom{0} & 66.7\phantom{0} & 83.9\phantom{0} & 92.4\phantom{0} \\
 & Slerp+TAT~\cite{jang2024sphericallinearinterpolationtextanchoring_slerp} & CC3M+LLaVA-Align+Laion-2M & 28.19 & 55.88 & 68.77 & 88.51 & 61.13 & 80.63 & 90.68 \\
 & CASE~\cite{levy2024data} & LaSCo & 30.89 & 60.75 & 73.88 & 92.84 & 60.17 & 80.17 & 90.41 \\
 & CASE~\cite{levy2024data} & LaSCo+COCO & 35.40 & 65.78 & 78.53 & 94.63 & 64.29 & 82.66 & 91.61 \\
 & ours & LaSCo+Hotel-CIR & \textbf{40.18} & \textbf{70.04} & \textbf{81.56} & \textbf{96.21} & \textbf{72.25} & \textbf{87.46} & \textbf{94.52} \\ \midrule
\multirow{9}{*}{ViT-L} & CompoDiff~\cite{gu2024compodiff} & ST18M+LAION2B & 22.35 & 54.36 & 73.41 & 91.77 & 62.55 & 81.44 & 90.21 \\
 & Pic2Word~\cite{saito2023pic2word} & CC3M & 23.90 & 51.70 & 65.30 & 87.80 & - & - & - \\
 & SEARLE-XL-OTI~\cite{baldrati2023zeroshot_searle} & CC3M & 24.87 & 52.31 & 66.29 & 88.58 & 53.80 & 74.31 & 86.94 \\
 & SEARLE-XL~\cite{baldrati2023zeroshot_searle} & CC3M & 24.24 & 52.48 & 66.29 & 88.84 & 53.76 & 75.01 & 88.19 \\
 & CIReVL~\cite{karthik2024visionbylanguage_cirevl} & - & 24.55 & 52.31 & 64.92 & 86.34 & 59.54 & 79.88 & 89.69 \\
 & LinCIR~\cite{gu2024lincir} & CC3M+SDP+COYO700M+OWT & 25.04 & 53.25 & 66.68 & - & 57.11 & 77.37 & 88.89 \\
 & LDRE~\cite{yang2024ldre} & - & 26.53 & 55.57 & 67.54 & 88.50 & 60.43 & 80.31 & 89.90 \\
 & Slerp+TAT~\cite{jang2024sphericallinearinterpolationtextanchoring_slerp} & CC3M+LLaVA-Align+Laion-2M & 30.94 & 59.40 & 70.94 & 89.18 & 64.70 & 82.92 & 92.31 \\
 & MagicLens~\cite{zhang2024magiclens} & web-scraped (36.7M) & 30.1\phantom{0} & 61.7\phantom{0} & 74.4\phantom{0} & 92.6\phantom{0} & 68.1\phantom{0} & 84.8\phantom{0} & 93.2\phantom{0} \\
 & CoVR-BLIP~\cite{ventura24covr} & WebVid-CoVR & 38.48 & 66.70 & 77.25 & 91.47 & 69.28 & 83.76 & 91.11 \\
 & ours & LaSCo+Hotel-CIR & \textbf{43.86} & \textbf{73.39} & \textbf{82.00} & \textbf{96.14} & \textbf{75.62} & \textbf{89.44} & \textbf{95.03} \\ \midrule
\multirow{2}{*}{ViT-H} 
  & LinCIR~\cite{gu2024lincir} & CC3M+SDP+COYO700M+OWT & 33.83 & 63.52 & 75.35 & - & 62.43 & 81.47 & 92.12  \\
 & ours & LaSCo+Hotel-CIR & \textbf{46.71} & \textbf{75.64} & \textbf{84.26} & \textbf{96.85} & \textbf{76.41} & \textbf{90.64} & \textbf{95.62} \\ \cdashline{1-10} \addlinespace[2pt]
\multirow{7}{*}{ViT-G}  & GRB+LCR~\cite{sun2024trainingfreezeroshotcomposedimage_grb_tfcir} & - & 30.92 & 56.99 & 68.58 & 85.28 & 66.67 & 78.68 & 82.60 \\
 & CompoDiff~\cite{gu2024compodiff} & ST18M+LAION2B & 32.39 & 57.61 & 77.25 & 94.61 & 67.88 & 85.29 & 94.07 \\
  & TFCIR~\cite{sun2024trainingfreezeroshotcomposedimage_grb_tfcir} & - & 32.82 & 61.13 & 71.76 & 85.28 & 66.63 & 78.58 & 82.68 \\
 & CIReVL~\cite{karthik2024visionbylanguage_cirevl} & - & 34.65 & 64.29 & 75.06 & 91.66 & 67.95 & 84.87 & 93.21 \\
 & LinCIR~\cite{gu2024lincir} & CC3M+SDP+COYO700M+OWT & 35.25 & 64.72 & 76.05 & - & 63.35 & 82.22 & 91.98 \\
 & LDRE~\cite{yang2024ldre} & - & 36.15 & 66.39 & 77.25 & 93.95 & 68.82 & 85.66 & 93.76 \\
 & CoVR-2~\cite{10685001_covr-2_blip} & CC-CoIR & 43.35 & 73.78 & 83.66 & 96.07 & 75.25 & 88.89 & 95.23 \\
 & CoVR-2~\cite{10685001_covr-2_blip} & WV-CC-CoVIR & 43.74 & 73.61 & 83.95 & 96.10 & 72.84 & 87.52 & 94.39 \\
 \bottomrule
\end{tabular}
\caption{Zero-shot composed image retrieval results on the CIRR test set. Higher is better for all metrics, best results are shown in bold.}
\label{tab:zero-shot_cirr}
\end{table*}

\begin{table}[h!]
\centering
\small
\begin{tabular}{@{}rlcccc@{}}
\toprule
 &  & \multicolumn{4}{c}{mAP@K} \\ \midrule
Size & \multicolumn{1}{l|}{Method} & K=5 & K=10 & K=25 & K=50 \\ \midrule
\multirow{7}{*}{B} & \multicolumn{1}{l|}{SEARLE~\cite{baldrati2023zeroshot_searle}} & \phantom{0}9.35 & \phantom{0}9.94 & 11.13 & 11.84 \\
 & \multicolumn{1}{l|}{CIReVL~\cite{karthik2024visionbylanguage_cirevl}} & 14.94 & 15.42 & 17.00 & 17.82 \\
 & \multicolumn{1}{l|}{LDRE~\cite{yang2024ldre}} & 17.96 & 18.32 & 20.21 & 21.11 \\
  & \multicolumn{1}{l|}{MagicLens~\cite{zhang2024magiclens}} & 23.1\phantom{0} & 23.8\phantom{0} & 25.8\phantom{0} & 26.7\phantom{0} \\
 & \multicolumn{1}{l|}{GRB+LCR~\cite{sun2024trainingfreezeroshotcomposedimage_grb_tfcir}} & 25.38 & 26.93 & 29.82 & 30.74 \\
 & \multicolumn{1}{l|}{TFCIR~\cite{sun2024trainingfreezeroshotcomposedimage_grb_tfcir}} & 26.52 & 28.25 & 31.23 & 31.99 \\
 & \multicolumn{1}{l|}{ours} & \textbf{28.12} & \textbf{29.42} & \textbf{32.26} & \textbf{33.54} \\ \midrule
\multirow{9}{*}{L} & \multicolumn{1}{l|}{SEARLE-XL~\cite{baldrati2023zeroshot_searle}} & 11.68 & 12.73 & 14.33 & 15.12 \\
 & \multicolumn{1}{l|}{CompoDiff~\cite{gu2024compodiff}} & 12.31 & 13.51 & 15.67 & 16.15 \\
 & \multicolumn{1}{l|}{LinCIR~\cite{gu2024lincir}} & 12.59 & 13.58 & 15.00 & 15.85 \\
 & \multicolumn{1}{l|}{CIReVL~\cite{karthik2024visionbylanguage_cirevl}} & 18.57 & 19.01 & 20.89 & 21.80 \\
 & \multicolumn{1}{l|}{CoVR-BLIP~\cite{ventura24covr}} & 21.43 & 22.33 & 24.47 & 25.48 \\
 & \multicolumn{1}{l|}{LDRE~\cite{yang2024ldre}} & 23.35 & 24.03 & 26.44 & 27.50 \\
  & \multicolumn{1}{l|}{MagicLens~\cite{zhang2024magiclens}} & 29.6\phantom{0} & 30.8\phantom{0} & 33.4\phantom{0} & 34.4\phantom{0} \\
 & \multicolumn{1}{l|}{ours} & \textbf{30.05} & \textbf{30.53} & \textbf{34.79} & \textbf{34.72} \\ \midrule
\multirow{1}{*}{H} 

  & \multicolumn{1}{l|}{ours} & \textbf{31.74} & \textbf{32.64} & \textbf{35.05} & \textbf{36.42} \\ \cdashline{1-6} \addlinespace[2pt]
  \multirow{5}{*}{G} 
& \multicolumn{1}{l|}{CompoDiff~\cite{gu2024compodiff} } & 15.33 & 17.71 & 19.45 & 21.01 \\
 & \multicolumn{1}{l|}{LinCIR~\cite{gu2024lincir} } & 19.71 & 21.01 & 23.13 & 24.18 \\
 & \multicolumn{1}{l|}{CIReVL~\cite{karthik2024visionbylanguage_cirevl} } & 26.77 & 27.59 & 29.96 & 31.03 \\
 & \multicolumn{1}{l|}{CoVR-2~\cite{10685001_covr-2_blip} } & 28.29 & 29.55 & 32.18 & 33.26 \\
& \multicolumn{1}{l|}{LDRE~\cite{yang2024ldre} } & 31.12 & 32.24 & 34.95 & 36.03 \\ 
\bottomrule
\end{tabular}
\caption{Zero-shot composed image retrieval results on the CIRCO test set. Higher is better for all metrics, best results are shown in bold.}
\label{tab:zero-shot_circo}
\end{table}

Table~\ref{tab:zero-shot_cirr} gives zero-shot (excluding CIRR from training data) composed image retrieval metrics on CIRR. {\name} demonstrates powerful zero-shot CIR performance across all encoder sizes, improving state-of-the-art R@1 by 4.78 for ViT-B, 5.38 for ViT-L, and 12.88 for ViT-H. Notably, our method with the ViT-H backbone also outperforms all methods using the much larger ViT-G. We observe that our method is able to outperform ViT-G-based CoVR-2~\cite{10685001_covr-2_blip} without pertaining on 4.9 million samples.

We also report zero-shot composed image retrieval performance on CIRCO~\cite{baldrati2023zeroshot_searle} in Table~\ref{tab:zero-shot_circo}. Our model outperforms all baseline methods when compared with the same backbone architecture, and outperforms all methods based on ViT-G, even as the ViT-G backbone has almost 400M additional parameters.

\subsection{Ablation Studies}
\begin{table}[h]
\small
\centering\begin{tabular}{@{}l|cccc@{}}
\toprule
 & \multicolumn{4}{c}{Recall@K} \\ \midrule
Data & K=1 & K=5 & K=10 & K=50 \\ \midrule
CIRR & 45.25 & 77.52 & 86.88 & 97.24 \\
CIRR+CIRR$_\mathbf{R}$ & 48.54 & 80.12 & 88.52 & 97.49 \\
CIRR+CIRR$_\mathbf{R}$+LaSCo & 53.12 & 83.78 & 89.48 & 98.67 \\
Aggregated & 55.24 & 84.85 & 90.75 & 98.82 \\ \bottomrule
\end{tabular}
\caption{Ablation study over dataset components w/ ViT-H backbone.}
\label{tab:ablation-data}
\end{table}
We evaluate {\name} on CIRR trained using different components of our Aggregated dataset in Table~\ref{tab:ablation-data}. We observe that {\name} performs better as more data is added. The significant performance increase from 45.25 to 48.54 R@1 from CIRR to CIRR+CIRR$_\mathbf{R}$ indicates that our data generation pipeline is highly effective at synthesizing data for training CIR models, producing large performance increases even without incorporating any additional imagery. Each subsequent large-scale dataset that does include new imagery (LaSCo and Hotel-CIR) gives an additional large performance boost. 

\begin{table}[h]
\centering
\small
\begin{tabular}{@{}l|cccc@{}}
\toprule
 & \multicolumn{4}{c}{Recall@K} \\ \midrule
Ablation & K=1 & K=5 & K=10 & K=50 \\ \midrule
Freeze text encoder & 49.52 & 78.46 & 86.33 & 94.25 \\
Freeze image encoder & 54.18 & 83.84 & 89.54 & 98.15 \\
No Query Negative & 54.52 & 83.95 & 89.57 & 98.12 \\
Only leaf NPs & 54.67 & 84.02 & 89.62 & 98.08 \\ 
No Text CC ($\lambda=0$) & 48.92 & 79.86 & 88.74 & 97.41 \\
With Text CC ($\lambda=0.08$) & 55.24 & 84.85 & 90.75 & 98.82 \\ \bottomrule %
\end{tabular}
\caption{Ablation study over model choices w/ ViT-H backbone.}
\label{tab:ablation-method}
\end{table}

\begin{figure*}[ht!]
    \centering
    \includegraphics[width=\textwidth]{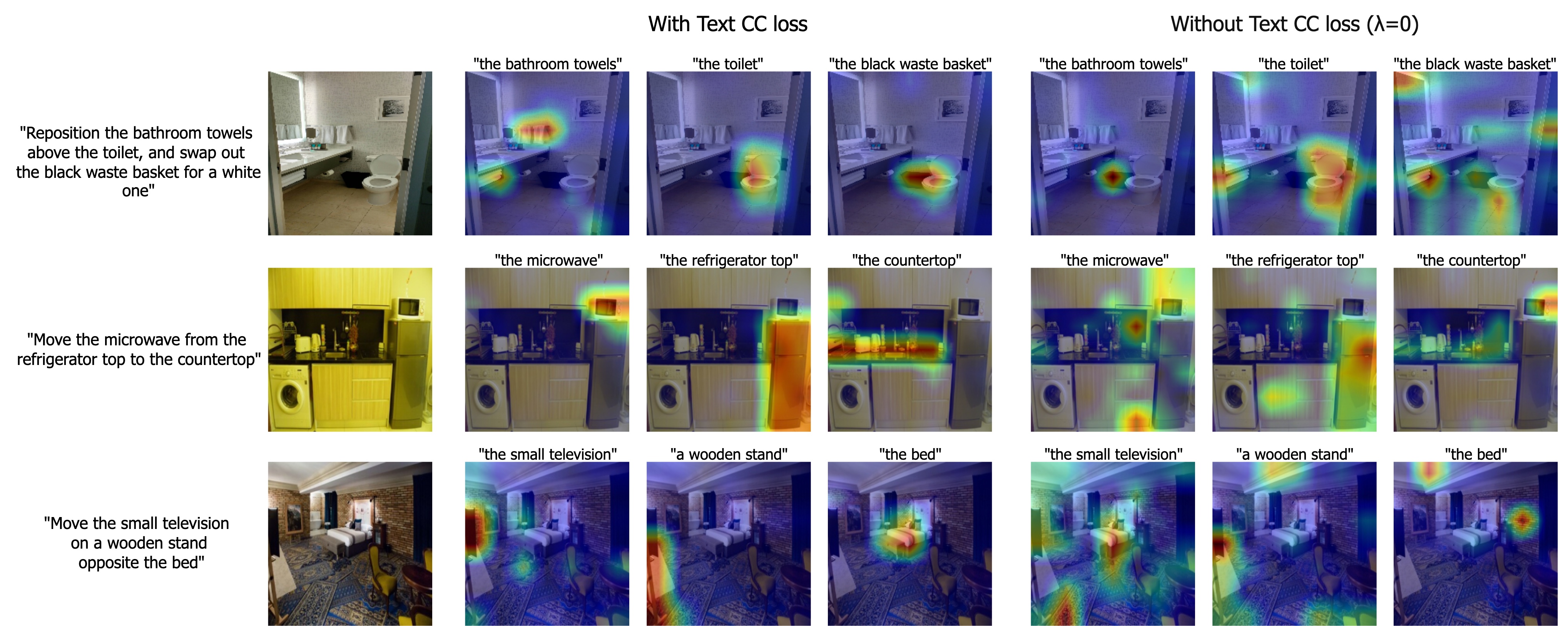}
    \caption{Noun-phrase-level cross attention maps for models trained with and without the Text Concept-Consistency loss.}
    \label{fig:xattn}
\end{figure*}

Next, we analyze the elements of {\name}'s design. We experiment with freezing the image or the text encoder, not considering branch-level noun phrases, and removing the Text CC loss. Freezing the text encoder results in a much larger performance drop when compared to freezing the image encoder (performance drop from 54.28 to 49.52 in R@1), indicating that the pre-trained CLIP text space is less conducive for CIR than the pre-trained CLIP image space. By comparison, only considering leaf noun phrases (a noun phrase that cannot be broken down into smaller syntactic units within the noun phrase itself) and removing the query image as a hard negative only results in a small decrease in performance. Finally, we see that removing the Text CC loss has a significant negative impact on performance (from 55.24 to 48.92 in R@1). This supports the idea that guiding the model to fuse representations of text concepts to their relevant image regions leads to higher performance. 

\subsection{Qualitative Results}
In Figure~\ref{fig:xattn}, we examine the cross attention of noun phrase representations to the image query for models trained with and without the Text CC loss. The attention maps are averaged across the cross-attention of the tokens \textit{from the representation of the whole text} in each noun phrase to the image query.

We qualitatively observe that the model's ability to ground the representations of noun phrases to their relevant image features improves significantly with the Text CC loss. Notably, the model trained without the Text CC loss often diffuses attention onto extraneous objects, such as the toilet and mirror in the first row. Furthermore, when the CC loss isn't used, the model often targets attention to completely wrong tokens. 
For instance, the model trained without the Text CC loss focuses attention associated with the concept ``the bathroom towels" onto the wastebasket. 
These observations demonstrate how the CC loss guides the model to learn to fuse concepts in text to their relevant image features \textit{without computing concept features at inference time}.

\section{Conclusions}

We present {\name}, a novel CIR method facilitating performant rapid retrieval over large image databases.  Our Text Concept-Consistency loss improves the concept-level alignment of the text and image embeddings and produces qualitatively more focused attention maps.  To train {\name}, we also introduce a data generation strategy that can be used to rewrite the annotations in existing CIR datasets or to generate new CIR datasets.  This method trained on these new datasets sets a new state-of-the-art in supervised and zero-shot composed image retrieval on multiple CIR benchmarks.

\clearpage
\setlength{\tabcolsep}{3.5 pt}

\maketitlesupplementary
\appendix

\begin{figure*}[h!]
    \centering
    \includegraphics[width=\linewidth]{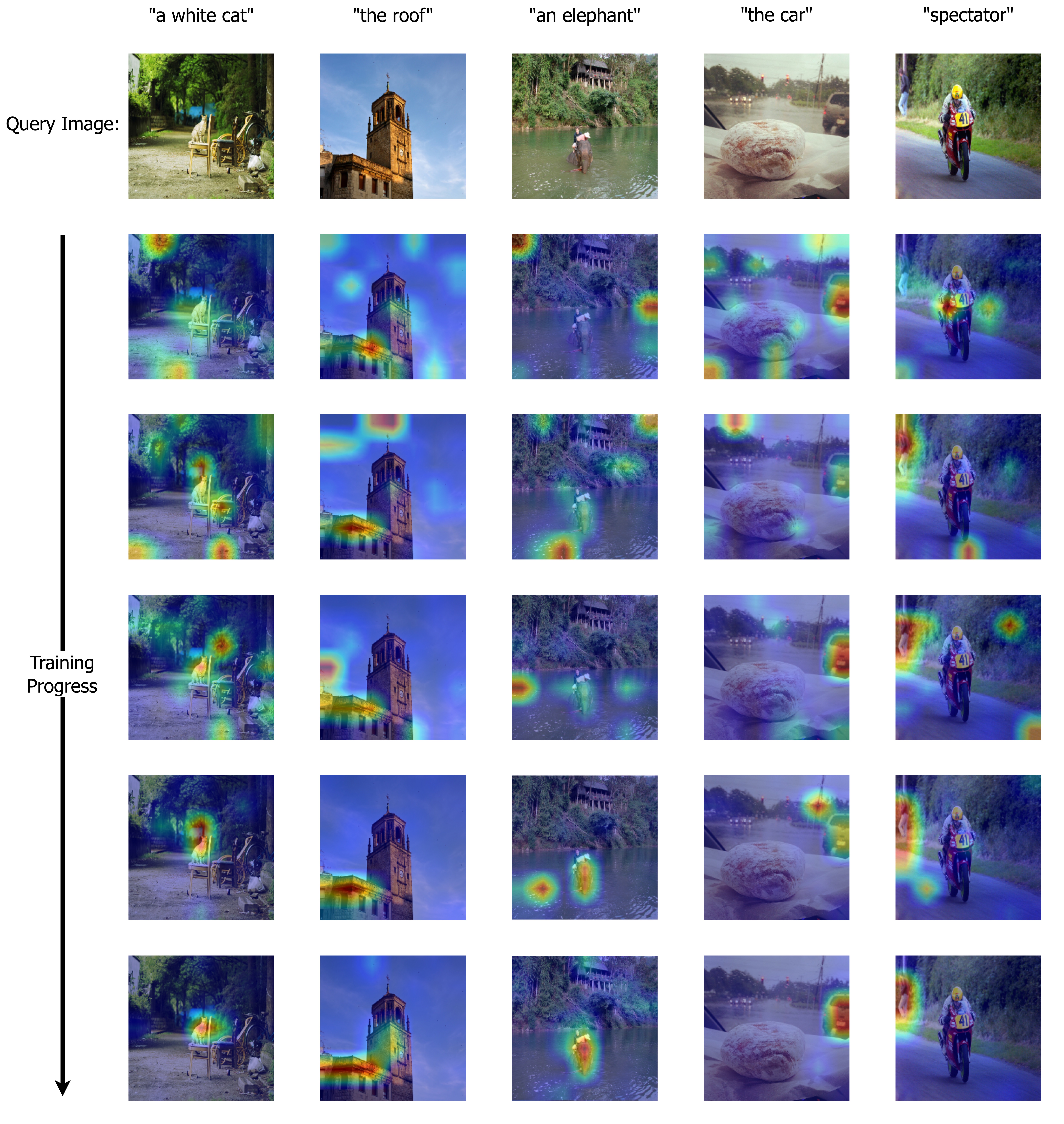}
    \caption{Visualization of attentions averaged over noun phrases to image queries. Concept attentions become more grounded to relevant image features with training.}
    \label{fig:training_progression}
\end{figure*}

\begin{figure*}[h!]
    \centering
    \includegraphics[width=\textwidth]{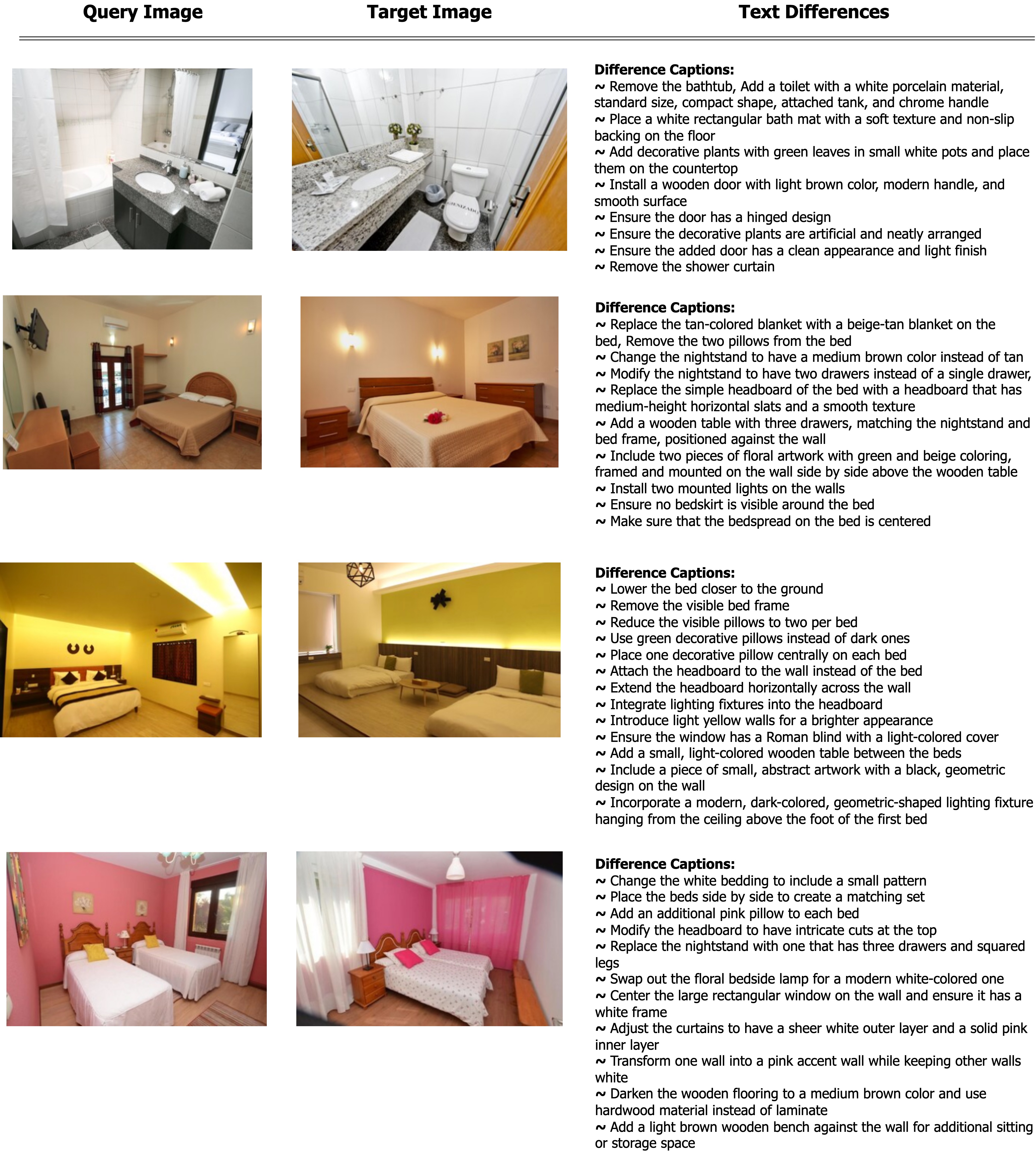}
    \caption{Example generated text differences for the Hotel-CIR dataset using our synthetic data generation pipeline.}
    \label{fig:hotels_examples}
\end{figure*}

\vfill
\begin{figure*}[h!]
    \centering
    \includegraphics[width=\textwidth]{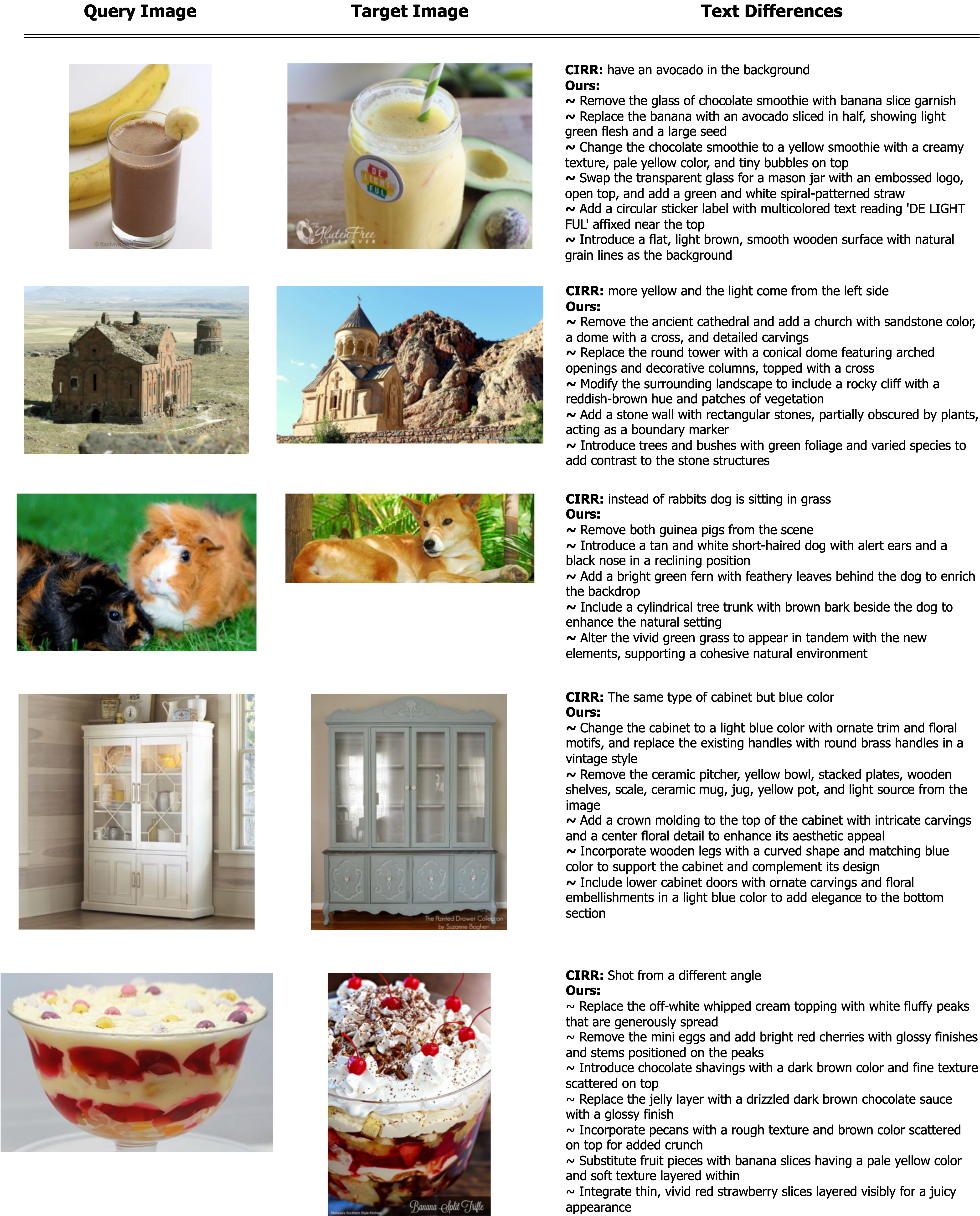}
    \caption{Example original reference texts from CIRR~\cite{Liu_2021_ICCV_cirplant} and generated text differences from the CIRR$_R$ dataset generated using our synthetic data generation pipeline.}
    \label{fig:cirr_examples}
\end{figure*}

\section*{Acknowledgements}
This material is based upon work done, in whole or in part, in coordination with the Department of Defense (DoD). Any opinions, findings, conclusions, or recommendations expressed in this material are those of the author(s) and do not necessarily reflect the views of the DoD and/or any agency or entity of the United States Government.

\section{Attention Visualization}
To demonstrate that optimizing {\name} for the Text Concept-Consistency and contrastive objectives improves the ability of the framework to ground text features to their relevant image features, we visualize the attention of concepts to the image query at various points in the training process. Figure~\ref{fig:training_progression} gives the concept attentions, averaged across noun phrases, for 5 concept-image pairs at 5 points during training (downwards is later in training). We observe that attention becomes significantly more focused to the image patches that text concepts refer to as training progresses, with very little spurious attention. In all cases, fully trained {\name} shows precise attentions to the image patches that a noun phrase refers to, while earlier models either ground attention to irrelevant objects or spurious image features.

\section{Synthetic Dataset Pipeline}
good4cir employs a three-stage pipeline to generate precise, high-quality annotations for CIR model training. In Stage 1, the model curates a list of key objects and descriptors from the query image. During Stage 2, the model derives a similar list from the target image by comparing it against the list of objects from the query image, ensuring consistency and making modifications when necessary. Stage 3 compares both lists to generate a list of fine-grained difference captions that describe the addition, removal, and modification of objects from the query to the target image. Figure ~\ref{fig:cirr_examples} and Figure ~\ref{fig:hotels_examples} give examples of synthesized examples from the good4cir framework. More details about the prompts used for generation and qualitative results validating the usefulness of good4cir's generated data for CIR models may be found in the accompanying paper ~\cite{kolouju2025good4cir}.

\section{Additional Benchmarks}
We also present zero-shot composed image retrieval metrics on FashionIQ~\cite{guo2019fashion} and the domain conversion on ImageNet-R~\cite{hendrycks2021manyfacesofrobustness, 5206848_IMAGENET} introduced by Pic2Word~\cite{saito2023pic2word}. The FashionIQ evaluation is stratified by classes of clothing items as defined by the dataset, and we report R@10 and R@50 on the test set. As described by Pic2Word, we use the 200 classes and domains outlined by ImageNet and ImageNet-R and perform domain level evaluation on retrieval results. The retrieval prompt is generated by picking a class from the set $\{ \texttt{cartoon, origami, toy,  sculpture} \}$. We report average R@10 and R@50 across target domains.

\begin{table}[h!]
\centering
\tiny
\begin{tabular}{@{}lcccccccc|cc@{}}
\toprule
\multirow{3}{*}{Method} & \multicolumn{8}{c|}{FashionIQ}                                                                                                        & \multicolumn{2}{c}{ImageNet-R}  \\ \cmidrule(l){2-11} 
                        & \multicolumn{2}{c}{Shirt}       & \multicolumn{2}{c}{Dress}       & \multicolumn{2}{c}{Toptee}      & \multicolumn{2}{c|}{Average}    & \multicolumn{2}{c}{Average}     \\ \cmidrule(l){2-11} 
                        & R@10           & R@50           & R@10           & R@50           & R@10           & R@50           & R@10           & R@50           & R@10           & R@50           \\ \midrule
LinCIR                  & 20.92          & 42.44          & 29.10          & 46.81          & 28.81          & 50.18          & 26.28          & 46.49          & -              & -              \\
iSEARLE-XL              & 28.75          & 47.84          & 22.51          & 46.36          & 31.31          & 52.68          & 27.52          & 48.96          & 15.52          & 33.39          \\
CIReVL                  & 29.49          & 47.40          & 24.79          & 44.76          & 31.36          & 53.65          & 28.55          & 48.57          & 23.75          & 43.05          \\
CoVR-BLIP-2             & 34.26          & 56.22          & 41.22          & 59.32          & 38.96          & 59.77          & 38.15          & 58.44          & -              & -              \\ \midrule
ours                    & \textbf{38.52} & \textbf{61.09} & \textbf{46.81} & \textbf{65.43} & \textbf{50.28} & \textbf{71.05} & \textbf{45.20} & \textbf{65.86} & \textbf{25.62} & \textbf{44.84} \\ \bottomrule
\end{tabular}
\caption{Zero-shot composed image retrieval metrics on FashionIQ and ImageNet-R.}
\label{tab:FIQ_Eval}
\end{table}

Table ~\ref{tab:FIQ_Eval} demonstrates that {\name} is consistently state-of-the-art, demonstrating convincing performance gains across both FashionIQ an ImageNet-R's domain translation task.  

\FloatBarrier
\clearpage
{
\small
\bibliographystyle{ieeenat_fullname}
\bibliography{main}

\begin{thebibliography}{74}
\providecommand{\natexlab}[1]{#1}
\providecommand{\url}[1]{\texttt{#1}}
\expandafter\ifx\csname urlstyle\endcsname\relax
  \providecommand{\doi}[1]{doi: #1}\else
  \providecommand{\doi}{doi: \begingroup \urlstyle{rm}\Url}\fi

\bibitem[mag()]{magiclens_github}
\url{https://github.com/google-deepmind/magiclens/issues/8}.

\bibitem[Bain et~al.(2021)Bain, Nagrani, Varol, and Zisserman]{Bain21_webvid}
Max Bain, Arsha Nagrani, G{\"u}l Varol, and Andrew Zisserman.
\newblock Frozen in time: A joint video and image encoder for end-to-end
  retrieval.
\newblock In \emph{IEEE International Conference on Computer Vision}, 2021.

\bibitem[Baldrati et~al.(2022)Baldrati, Bertini, Uricchio, and
  Del~Bimbo]{lfclip_lfblip_9879245}
Alberto Baldrati, Marco Bertini, Tiberio Uricchio, and Alberto Del~Bimbo.
\newblock Effective conditioned and composed image retrieval combining
  clip-based features.
\newblock In \emph{2022 IEEE/CVF Conference on Computer Vision and Pattern
  Recognition (CVPR)}, pages 21434--21442, 2022.

\bibitem[Baldrati et~al.(2023{\natexlab{a}})Baldrati, Agnolucci, Bertini, and
  Del~Bimbo]{baldrati2023zeroshot_searle}
Alberto Baldrati, Lorenzo Agnolucci, Marco Bertini, and Alberto Del~Bimbo.
\newblock Zero-shot composed image retrieval with textual inversion.
\newblock In \emph{Proceedings of the IEEE/CVF International Conference on
  Computer Vision (ICCV)}, pages 15338--15347, 2023{\natexlab{a}}.

\bibitem[Baldrati et~al.(2023{\natexlab{b}})Baldrati, Bertini, Uricchio, and
  Bimbo]{baldrati2023composed_clip4cir}
Alberto Baldrati, Marco Bertini, Tiberio Uricchio, and Alberto~Del Bimbo.
\newblock Composed image retrieval using contrastive learning and task-oriented
  clip-based features.
\newblock \emph{ACM Transactions on Multimedia Computing, Communications and
  Applications}, 20\penalty0 (3):\penalty0 1--24, 2023{\natexlab{b}}.

\bibitem[Brooks et~al.(2022)Brooks, Holynski, and
  Efros]{brooks2022instructpix2pix}
Tim Brooks, Aleksander Holynski, and Alexei~A Efros.
\newblock Instructpix2pix: Learning to follow image editing instructions.
\newblock \emph{arXiv preprint arXiv:2211.09800}, 2022.

\bibitem[Caron et~al.(2021)Caron, Touvron, Misra, J\'egou, Mairal, Bojanowski,
  and Joulin]{caron2021emerging}
Mathilde Caron, Hugo Touvron, Ishan Misra, Herv\'e J\'egou, Julien Mairal,
  Piotr Bojanowski, and Armand Joulin.
\newblock Emerging properties in self-supervised vision transformers.
\newblock In \emph{ICCV}, 2021.

\bibitem[Chen et~al.(2023{\natexlab{a}})Chen, Liu, Wang, Bakker, Georgiou,
  Fieguth, Liu, and Lew]{survey}
Wei Chen, Yu Liu, Weiping Wang, Erwin~M. Bakker, Theodoros Georgiou, Paul
  Fieguth, Li Liu, and Michael~S. Lew.
\newblock Deep learning for instance retrieval: A survey.
\newblock \emph{IEEE Transactions on Pattern Analysis and Machine
  Intelligence}, 45\penalty0 (6):\penalty0 7270--7292, 2023{\natexlab{a}}.

\bibitem[Chen et~al.(2020)Chen, Gong, and
  Bazzani]{9157634_vislingattnlearningchen}
Yanbei Chen, Shaogang Gong, and Loris Bazzani.
\newblock Image search with text feedback by visiolinguistic attention
  learning.
\newblock In \emph{IEEE/CVF Conference on Computer Vision and Pattern
  Recognition (CVPR)}, pages 2998--3008, 2020.

\bibitem[Chen et~al.(2023{\natexlab{b}})Chen, Yuan, Tian, Geng, Li, Zhou,
  Metaxas, and Yang]{inproceedings_fdtpaper}
Yuxiao Chen, Jianbo Yuan, Yu Tian, Shijie Geng, Xinyu Li, Ding Zhou,
  Dimitris~N. Metaxas, and Hongxia Yang.
\newblock Revisiting multimodal representation in contrastive learning: From
  patch and token embeddings to finite discrete tokens.
\newblock In \emph{Proceedings of the IEEE/CVF Conference on Computer Vision
  and Pattern Recognition (CVPR)}, pages 15095--15104, 2023{\natexlab{b}}.

\bibitem[Dai et~al.(2024)Dai, Li, Li, Tiong, Zhao, Wang, Li, Fung, and
  Hoi]{dai2023instructblipgeneralpurposevisionlanguagemodels}
Wenliang Dai, Junnan Li, Dongxu Li, Anthony Meng~Huat Tiong, Junqi Zhao,
  Weisheng Wang, Boyang Li, Pascale Fung, and Steven Hoi.
\newblock Instructblip: towards general-purpose vision-language models with
  instruction tuning.
\newblock In \emph{Proceedings of the 37th International Conference on Neural
  Information Processing Systems}, Red Hook, NY, USA, 2024. Curran Associates
  Inc.

\bibitem[Delmas et~al.(2022)Delmas, Rezende, Csurka, and
  Larlus]{delmas2022artemis}
Ginger Delmas, Rafael~S Rezende, Gabriela Csurka, and Diane Larlus.
\newblock Artemis: Attention-based retrieval with text-explicit matching and
  implicit similarity.
\newblock In \emph{International Conference on Learning Representations}, 2022.

\bibitem[Deng et~al.(2009)Deng, Dong, Socher, Li, Li, and
  Fei-Fei]{5206848_IMAGENET}
Jia Deng, Wei Dong, Richard Socher, Li-Jia Li, Kai Li, and Li Fei-Fei.
\newblock Imagenet: A large-scale hierarchical image database.
\newblock In \emph{2009 IEEE Conference on Computer Vision and Pattern
  Recognition}, pages 248--255, 2009.

\bibitem[Flemings et~al.(2024)Flemings, Zhang, Jiang, Takhirov, and
  Annavaram]{hallucination}
James Flemings, Wanrong Zhang, Bo Jiang, Zafar Takhirov, and Murali Annavaram.
\newblock Characterizing context influence and hallucination in summarization,
  2024.

\bibitem[Forbes et~al.(2019)Forbes, Kaeser-Chen, Sharma, and
  Belongie]{forbes2019neural}
Maxwell Forbes, Christine Kaeser-Chen, Piyush Sharma, and Serge Belongie.
\newblock Neural naturalist: Generating fine-grained image comparisons.
\newblock \emph{arXiv preprint arXiv:1909.04101}, 2019.

\bibitem[Goyal et~al.(2017)Goyal, Khot, Summers{-}Stay, Batra, and
  Parikh]{balanced_vqa_v2}
Yash Goyal, Tejas Khot, Douglas Summers{-}Stay, Dhruv Batra, and Devi Parikh.
\newblock Making the {V} in {VQA} matter: Elevating the role of image
  understanding in {V}isual {Q}uestion {A}nswering.
\newblock In \emph{Conference on Computer Vision and Pattern Recognition
  (CVPR)}, 2017.

\bibitem[Gu et~al.(2024{\natexlab{a}})Gu, Chun, Kim, , Kang, and
  Yun]{gu2024lincir}
Geonmo Gu, Sanghyuk Chun, Wonjae Kim, , Yoohoon Kang, and Sangdoo Yun.
\newblock Language-only training of zero-shot composed image retrieval.
\newblock In \emph{Conference on Computer Vision and Pattern Recognition
  (CVPR)}, 2024{\natexlab{a}}.

\bibitem[Gu et~al.(2024{\natexlab{b}})Gu, Chun, Kim, Jun, Kang, and
  Yun]{gu2024compodiff}
Geonmo Gu, Sanghyuk Chun, Wonjae Kim, HeeJae Jun, Yoohoon Kang, and Sangdoo
  Yun.
\newblock Compodiff: Versatile composed image retrieval with latent diffusion.
\newblock \emph{Transactions on Machine Learning Research}, 2024{\natexlab{b}}.
\newblock Expert Certification.

\bibitem[Hendrycks et~al.(2021)Hendrycks, Basart, Mu, Kadavath, Wang, Dorundo,
  Desai, Zhu, Parajuli, Guo, Song, Steinhardt, and
  Gilmer]{hendrycks2021manyfacesofrobustness}
Dan Hendrycks, Steven Basart, Norman Mu, Saurav Kadavath, Frank Wang, Evan
  Dorundo, Rahul Desai, Tyler Zhu, Samyak Parajuli, Mike Guo, Dawn Song, Jacob
  Steinhardt, and Justin Gilmer.
\newblock The many faces of robustness: A critical analysis of
  out-of-distribution generalization.
\newblock \emph{ICCV}, 2021.

\bibitem[Hertz et~al.(2022)Hertz, Mokady, Tenenbaum, Aberman, Pritch, and
  Cohen-Or]{hertz2022prompt}
Amir Hertz, Ron Mokady, Jay Tenenbaum, Kfir Aberman, Yael Pritch, and Daniel
  Cohen-Or.
\newblock Prompt-to-prompt image editing with cross attention control.
\newblock 2022.

\bibitem[Huang et~al.(2020)Huang, Hui, Liu, Li, Wei, Han, Liu, and
  Li]{huang2020referring}
Shaofei Huang, Tianrui Hui, Si Liu, Guanbin Li, Yunchao Wei, Jizhong Han, Luoqi
  Liu, and Bo Li.
\newblock Referring image segmentation via cross-modal progressive
  comprehension.
\newblock In \emph{Proceedings of the IEEE/CVF Conference on Computer Vision
  and Pattern Recognition}, pages 10488--10497, 2020.

\bibitem[Hudson and Manning(2019)]{hudson2019gqa}
Drew~A Hudson and Christopher~D Manning.
\newblock Gqa: A new dataset for real-world visual reasoning and compositional
  question answering.
\newblock In \emph{Proceedings of the IEEE/CVF conference on computer vision
  and pattern recognition}, pages 6700--6709, 2019.

\bibitem[Ilharco et~al.(2021)Ilharco, Wortsman, Wightman, Gordon, Carlini,
  Taori, Dave, Shankar, Namkoong, Miller, Hajishirzi, Farhadi, and
  Schmidt]{ilharco_gabriel_2021_5143773_openclip}
Gabriel Ilharco, Mitchell Wortsman, Ross Wightman, Cade Gordon, Nicholas
  Carlini, Rohan Taori, Achal Dave, Vaishaal Shankar, Hongseok Namkoong, John
  Miller, Hannaneh Hajishirzi, Ali Farhadi, and Ludwig Schmidt.
\newblock Openclip, 2021.
\newblock If you use this software, please cite it as below.

\bibitem[Jang et~al.(2024)Jang, Huynh, Shah, Chen, and
  Lim]{jang2024sphericallinearinterpolationtextanchoring_slerp}
Young~Kyun Jang, Dat Huynh, Ashish Shah, Wen-Kai Chen, and Ser-Nam Lim.
\newblock Spherical linear interpolation and text-anchoring for zero-shot
  composed image retrieval.
\newblock arXiv preprint arXiv:2405.00571, 2024.

\bibitem[Jiang et~al.(2024)Jiang, Wang, Wu, Wang, and
  Qian]{jiang2024dualrelationalignmentcomposed}
Xintong Jiang, Yaxiong Wang, Yujiao Wu, Meng Wang, and Xueming Qian.
\newblock Dual relation alignment for composed image retrieval, 2024.

\bibitem[Johnson et~al.(2017)Johnson, Hariharan, van~der Maaten, Fei-Fei,
  Zitnick, and Girshick]{johnson2017clevr}
Justin Johnson, Bharath Hariharan, Laurens van~der Maaten, Li Fei-Fei,
  C~Lawrence Zitnick, and Ross Girshick.
\newblock Clevr: A diagnostic dataset for compositional language and elementary
  visual reasoning.
\newblock In \emph{CVPR}, 2017.

\bibitem[Karthik et~al.(2024)Karthik, Roth, Mancini, and
  Akata]{karthik2024visionbylanguage_cirevl}
Shyamgopal Karthik, Karsten Roth, Massimiliano Mancini, and Zeynep Akata.
\newblock Vision-by-language for training-free compositional image retrieval.
\newblock \emph{International Conference on Learning Representations (ICLR)},
  2024.

\bibitem[Kirillov et~al.(2023)Kirillov, Mintun, Ravi, Mao, Rolland, Gustafson,
  Xiao, Whitehead, Berg, Lo, Doll{\'a}r, and Girshick]{kirillov2023segany}
Alexander Kirillov, Eric Mintun, Nikhila Ravi, Hanzi Mao, Chloe Rolland, Laura
  Gustafson, Tete Xiao, Spencer Whitehead, Alexander~C. Berg, Wan-Yen Lo, Piotr
  Doll{\'a}r, and Ross Girshick.
\newblock Segment anything.
\newblock \emph{arXiv:2304.02643}, 2023.

\bibitem[Kolouju et~al.(2025)Kolouju, Xing, Pless, Jacobs, and
  Stylianou]{kolouju2025good4cir}
Pranavi Kolouju, Eric Xing, Robert Pless, Nathan Jacobs, and Abby Stylianou.
\newblock good4cir: Generating detailed synthetic captions for composed image
  retrieval, 2025.

\bibitem[Lee et~al.(2019)Lee, Lee, Kim, Kosiorek, Choi, and
  Teh]{lee2019set_attnpooling}
Juho Lee, Yoonho Lee, Jungtaek Kim, Adam Kosiorek, Seungjin Choi, and Yee~Whye
  Teh.
\newblock Set transformer: A framework for attention-based
  permutation-invariant neural networks.
\newblock In \emph{Proceedings of the 36th International Conference on Machine
  Learning}, pages 3744--3753, 2019.

\bibitem[Levy et~al.(2024)Levy, Ben-Ari, Darshan, and Lischinski]{levy2024data}
M. Levy, R. Ben-Ari, N. Darshan, and D. Lischinski.
\newblock Data roaming and quality assessment for composed image retrieval.
\newblock In \emph{AAAI}, 2024.

\bibitem[Li et~al.(2022{\natexlab{a}})Li, Weinberger, Belongie, Koltun, and
  Ranftl]{li2022languagedriven}
Boyi Li, Kilian~Q Weinberger, Serge Belongie, Vladlen Koltun, and Rene Ranftl.
\newblock Language-driven semantic segmentation.
\newblock In \emph{ICLR}, 2022{\natexlab{a}}.

\bibitem[Li and Zhu(2023)]{10219821_vl_alignment}
Dafeng Li and Yingying Zhu.
\newblock Visual-linguistic alignment and composition for image retrieval with
  text feedback.
\newblock In \emph{2023 IEEE International Conference on Multimedia and Expo
  (ICME)}, pages 108--113, 2023.

\bibitem[Li et~al.(2022{\natexlab{b}})Li, Li, Xiong, and Hoi]{li2022blip}
Junnan Li, Dongxu Li, Caiming Xiong, and Steven Hoi.
\newblock Blip: Bootstrapping language-image pre-training for unified
  vision-language understanding and generation.
\newblock In \emph{International conference on machine learning}, pages
  12888--12900. PMLR, 2022{\natexlab{b}}.

\bibitem[Li* et~al.(2022)Li*, Zhang*, Zhang*, Yang, Li, Zhong, Wang, Yuan,
  Zhang, Hwang, Chang, and Gao]{li2021groundedglip}
Liunian~Harold Li*, Pengchuan Zhang*, Haotian Zhang*, Jianwei Yang, Chunyuan
  Li, Yiwu Zhong, Lijuan Wang, Lu Yuan, Lei Zhang, Jenq-Neng Hwang, Kai-Wei
  Chang, and Jianfeng Gao.
\newblock Grounded language-image pre-training.
\newblock In \emph{CVPR}, 2022.

\bibitem[Lin et~al.(2014)Lin, Maire, Belongie, Hays, Perona, Ramanan,
  Doll{\'a}r, and Zitnick]{lin2015microsoftcococommonobjects}
Tsung-Yi Lin, Michael Maire, Serge Belongie, James Hays, Pietro Perona, Deva
  Ramanan, Piotr Doll{\'a}r, and C.~Lawrence Zitnick.
\newblock Microsoft coco: Common objects in context.
\newblock In \emph{Computer Vision -- ECCV 2014}, pages 740--755, Cham, 2014.
  Springer International Publishing.

\bibitem[Liu et~al.(2023{\natexlab{a}})Liu, Li, Wu, and
  Lee]{liu2023visualinstructiontuning_llava}
Haotian Liu, Chunyuan Li, Qingyang Wu, and Yong~Jae Lee.
\newblock Visual instruction tuning.
\newblock In \emph{Advances in Neural Information Processing Systems}, pages
  34892--34916. Curran Associates, Inc., 2023{\natexlab{a}}.

\bibitem[Liu et~al.(2023{\natexlab{b}})Liu, Zeng, Ren, Li, Zhang, Yang, Jiang,
  Li, Yang, Su, et~al.]{liu2023grounding}
Shilong Liu, Zhaoyang Zeng, Tianhe Ren, Feng Li, Hao Zhang, Jie Yang, Qing
  Jiang, Chunyuan Li, Jianwei Yang, Hang Su, et~al.
\newblock Grounding dino: Marrying dino with grounded pre-training for open-set
  object detection.
\newblock \emph{arXiv preprint arXiv:2303.05499}, 2023{\natexlab{b}}.

\bibitem[Liu et~al.(2021)Liu, Rodriguez-Opazo, Teney, and
  Gould]{Liu_2021_ICCV_cirplant}
Zheyuan Liu, Cristian Rodriguez-Opazo, Damien Teney, and Stephen Gould.
\newblock Image retrieval on real-life images with pre-trained
  vision-and-language models.
\newblock In \emph{Proceedings of the IEEE/CVF International Conference on
  Computer Vision (ICCV)}, pages 2125--2134, 2021.

\bibitem[Long et~al.(2024)Long, Ge, Mccreadie, and
  Jose]{long2024cfirfasteffectivelongtext}
Zijun Long, Xuri Ge, Richard Mccreadie, and Joemon Jose.
\newblock Cfir: Fast and effective long-text to image retrieval for large
  corpora, 2024.

\bibitem[Loshchilov and
  Hutter(2019)]{loshchilov2019decoupledweightdecayregularization}
Ilya Loshchilov and Frank Hutter.
\newblock Decoupled weight decay regularization.
\newblock In \emph{International Conference on Learning Representations}, 2019.

\bibitem[Murphy(1990)]{MURPHY1990259_nounphrase}
G. Murphy.
\newblock Noun phrase interpretation and conceptual combination.
\newblock \emph{Journal of Memory and Language}, 29\penalty0 (3), 1990.

\bibitem[Psomas et~al.(2024)Psomas, Kakogeorgiou, Efthymiadis, Tolias, Chum,
  Avrithis, and Karantzalos]{psomas2024composed}
B. Psomas, I. Kakogeorgiou, N. Efthymiadis, G. Tolias, O. Chum, Y. Avrithis,
  and K. Karantzalos.
\newblock Composed image retrieval for remote sensing.
\newblock In \emph{IGARSS 2024 - 2024 IEEE International Geoscience and Remote
  Sensing Symposium}, 2024.

\bibitem[Qi et~al.(2020)Qi, Zhang, Zhang, Bolton, and Manning]{qi2020stanza}
Peng Qi, Yuhao Zhang, Yuhui Zhang, Jason Bolton, and Christopher~D. Manning.
\newblock Stanza: A {Python} natural language processing toolkit for many human
  languages.
\newblock In \emph{Proceedings of the 58th Annual Meeting of the Association
  for Computational Linguistics: System Demonstrations}, 2020.

\bibitem[Radford et~al.(2021{\natexlab{a}})Radford, Kim, Hallacy, Ramesh, Goh,
  Agarwal, Sastry, Askell, Mishkin, Clark, Krueger, and
  Sutskever]{pmlr-v139-radford21a}
Alec Radford, Jong~Wook Kim, Chris Hallacy, Aditya Ramesh, Gabriel Goh,
  Sandhini Agarwal, Girish Sastry, Amanda Askell, Pamela Mishkin, Jack Clark,
  Gretchen Krueger, and Ilya Sutskever.
\newblock Learning transferable visual models from natural language
  supervision.
\newblock In \emph{Proceedings of the 38th International Conference on Machine
  Learning}, pages 8748--8763. PMLR, 2021{\natexlab{a}}.

\bibitem[Radford et~al.(2021{\natexlab{b}})Radford, Kim, Hallacy, Ramesh, Goh,
  Agarwal, Sastry, Askell, Mishkin, Clark, et~al.]{radford2021learning}
Alec Radford, Jong~Wook Kim, Chris Hallacy, Aditya Ramesh, Gabriel Goh,
  Sandhini Agarwal, Girish Sastry, Amanda Askell, Pamela Mishkin, Jack Clark,
  et~al.
\newblock Learning transferable visual models from natural language
  supervision.
\newblock In \emph{International conference on machine learning}, pages
  8748--8763. PMLR, 2021{\natexlab{b}}.

\bibitem[Ren et~al.(2024)Ren, Liu, Zeng, Lin, Li, Cao, Chen, Huang, Chen, Yan,
  Zeng, Zhang, Li, Yang, Li, Jiang, and Zhang]{ren2024grounded}
Tianhe Ren, Shilong Liu, Ailing Zeng, Jing Lin, Kunchang Li, He Cao, Jiayu
  Chen, Xinyu Huang, Yukang Chen, Feng Yan, Zhaoyang Zeng, Hao Zhang, Feng Li,
  Jie Yang, Hongyang Li, Qing Jiang, and Lei Zhang.
\newblock Grounded sam: Assembling open-world models for diverse visual tasks,
  2024.

\bibitem[Saito et~al.(2023)Saito, Sohn, Zhang, Li, Lee, Saenko, and
  Pfister]{saito2023pic2word}
Kuniaki Saito, Kihyuk Sohn, Xiang Zhang, Chun-Liang Li, Chen-Yu Lee, Kate
  Saenko, and Tomas Pfister.
\newblock Pic2word: Mapping pictures to words for zero-shot composed image
  retrieval.
\newblock \emph{CVPR}, 2023.

\bibitem[Shankar et~al.(2017)Shankar, Narumanchi, Ananya, Kompalli, and
  Chaudhury]{shankar2017deeplearningbasedlarge}
Devashish Shankar, Sujay Narumanchi, H~A Ananya, Pramod Kompalli, and
  Krishnendu Chaudhury.
\newblock Deep learning based large scale visual recommendation and search for
  e-commerce, 2017.

\bibitem[Simsar et~al.(2023)Simsar, Tonioni, Xian, Hofmann, and
  Tombari]{simsar2023limelocalizedimageediting}
Enis Simsar, Alessio Tonioni, Yongqin Xian, Thomas Hofmann, and Federico
  Tombari.
\newblock Lime: Localized image editing via attention regularization in
  diffusion models, 2023.

\bibitem[Song et~al.(2024)Song, Hwang, Yoon, Choi, and Gu]{song2024syncmask}
Chull~Hwan Song, Taebaek Hwang, Jooyoung Yoon, Shunghyun Choi, and Yeong~Hyeon
  Gu.
\newblock Syncmask: Synchronized attentional masking for fashion-centric
  vision-language pretraining.
\newblock In \emph{Proceedings of the IEEE/CVF Conference on Computer Vision
  and Pattern Recognition}, pages 13948--13957, 2024.

\bibitem[Song et~al.(2023)Song, Ma, Zou, Zhang, and Huang]{song2023FDalign}
Kun Song, Huimin Ma, Bochao Zou, Huishuai Zhang, and Weiran Huang.
\newblock Fd-align: Feature discrimination alignment for fine-tuning
  pre-trained models in few-shot learning.
\newblock \emph{NeurIPS}, 2023.

\bibitem[Stylianou et~al.(2019)Stylianou, Xuan, Shende, Brandt, Souvenir, and
  Pless]{hotels50k}
Abby Stylianou, Hong Xuan, Maya Shende, Jonathan Brandt, Richard Souvenir, and
  Robert Pless.
\newblock Hotels-50k: A global hotel recognition dataset.
\newblock In \emph{The AAAI Conference on Artificial Intelligence (AAAI)},
  2019.

\bibitem[Sudhish et~al.(2024)Sudhish, Nair, and S]{SUDHISH2024105620}
Dhanya~K. Sudhish, Latha~R. Nair, and Shailesh S.
\newblock Content-based image retrieval for medical diagnosis using fuzzy
  clustering and deep learning.
\newblock \emph{Biomedical Signal Processing and Control}, 88:\penalty0 105620,
  2024.

\bibitem[Suhr et~al.(2017)Suhr, Lewis, Yeh, and Artzi]{suhr2017corpus}
Alane Suhr, Mike Lewis, James Yeh, and Yoav Artzi.
\newblock A corpus of natural language for visual reasoning.
\newblock In \emph{Proceedings of the 55th Annual Meeting of the Association
  for Computational Linguistics (Volume 2: Short Papers)}, pages 217--223,
  2017.

\bibitem[Suhr et~al.(2019)Suhr, Zhou, Zhang, Zhang, Bai, and
  Artzi]{suhr2018corpus}
Alane Suhr, Stephanie Zhou, Ally Zhang, Iris Zhang, Huajun Bai, and Yoav Artzi.
\newblock A corpus for reasoning about natural language grounded in
  photographs.
\newblock 1:\penalty0 6418--6428, 2019.

\bibitem[Sultan et~al.(2023)Sultan, Jacobs, Stylianou, and
  Pless]{sultan2023exploring}
Manal Sultan, Lia Jacobs, Abby Stylianou, and Robert Pless.
\newblock Exploring clip for real world, text-based image retrieval.
\newblock In \emph{2023 IEEE Applied Imagery Pattern Recognition Workshop
  (AIPR)}, pages 1--6, 2023.

\bibitem[Sun et~al.(2024)Sun, Ye, and
  Gong]{sun2024trainingfreezeroshotcomposedimage_grb_tfcir}
Shitong Sun, Fanghua Ye, and Shaogang Gong.
\newblock Training-free zero-shot composed image retrieval with local concept
  reranking.
\newblock arXiv preprint arXiv:2312.08924, 2024.

\bibitem[Ventura et~al.(2024{\natexlab{a}})Ventura, Yang, Schmid, and
  Varol]{10685001_covr-2_blip}
L. Ventura, A. Yang, C. Schmid, and G. Varol.
\newblock { CoVR-2: Automatic Data Construction for Composed Video Retrieval }.
\newblock \emph{IEEE Transactions on Pattern Analysis \& Machine Intelligence},
  46\penalty0 (12):\penalty0 11409--11421, 2024{\natexlab{a}}.

\bibitem[Ventura et~al.(2024{\natexlab{b}})Ventura, Yang, Schmid, and
  Varol]{ventura24covr}
L. Ventura, A. Yang, C. Schmid, and G. Varol.
\newblock {CoVR}: Learning composed video retrieval from web video captions.
\newblock \emph{AAAI}, 2024{\natexlab{b}}.

\bibitem[Ventura et~al.(2024{\natexlab{c}})Ventura, Yang, Schmid, and
  Varol]{ventura24covr2}
L. Ventura, A. Yang, C. Schmid, and G. Varol.
\newblock {CoVR-2}: Automatic data construction for composed video retrieval.
\newblock \emph{IEEE TPAMI}, 2024{\natexlab{c}}.

\bibitem[Vo et~al.(2019)Vo, Jiang, Sun, Murphy, Li, Fei-Fei, and
  Hays]{vo2019composingempiricalodyssey}
Nam Vo, Lu Jiang, Chen Sun, Kevin Murphy, Li-Jia Li, Li Fei-Fei, and James
  Hays.
\newblock Composing text and image for image retrieval-an empirical odyssey.
\newblock In \emph{Proceedings of the IEEE/CVF conference on computer vision
  and pattern recognition}, pages 6439--6448, 2019.

\bibitem[Wan et~al.(2025)Wan, Cho, Stengel-Eskin, and Bansal]{CRG}
David Wan, Jaemin Cho, Elias Stengel-Eskin, and Mohit Bansal.
\newblock Contrastive region guidance: Improving grounding in vision-language
  models without training.
\newblock In \emph{ECCV}, 2025.

\bibitem[Wan et~al.(2024)Wan, Wang, Zou, and Zhang]{Wan_2024_CVPR}
Yongquan Wan, Wenhai Wang, Guobing Zou, and Bofeng Zhang.
\newblock Cross-modal feature alignment and fusion for composed image
  retrieval.
\newblock In \emph{Proceedings of the IEEE/CVF Conference on Computer Vision
  and Pattern Recognition (CVPR) Workshops}, pages 8384--8388, 2024.

\bibitem[Wang et~al.(2023)Wang, Yang, Yang, Butt, and van~de
  Weijer]{NEURIPS2023_5321b1da}
Kai Wang, Fei Yang, Shiqi Yang, Muhammad~Atif Butt, and Joost van~de Weijer.
\newblock Dynamic prompt learning: Addressing cross-attention leakage for
  text-based image editing.
\newblock In \emph{Advances in Neural Information Processing Systems}, pages
  26291--26303. Curran Associates, Inc., 2023.

\bibitem[Wen et~al.(2023)Wen, Zhang, Song, Wei, and Nie]{Wen_2023}
Haokun Wen, Xian Zhang, Xuemeng Song, Yinwei Wei, and Liqiang Nie.
\newblock Target-guided composed image retrieval.
\newblock In \emph{Proceedings of the 31st ACM International Conference on
  Multimedia}, page 915–923. ACM, 2023.

\bibitem[Wu et~al.(2021)Wu, Gao, Guo, Al-Halah, Rennie, Grauman, and
  Feris]{guo2019fashion}
Hui Wu, Yupeng Gao, Xiaoxiao Guo, Ziad Al-Halah, Steven Rennie, Kristen
  Grauman, and Rogerio Feris.
\newblock The fashion iq dataset: Retrieving images by combining side
  information and relative natural language feedback.
\newblock \emph{CVPR}, 2021.

\bibitem[Wu et~al.(2023)Wu, Wang, Zhou, Lu, and Li]{asymmetric}
Hui Wu, Min Wang, Wengang Zhou, Zhenbo Lu, and Houqiang Li.
\newblock { Asymmetric Feature Fusion for Image Retrieval }.
\newblock In \emph{2023 IEEE/CVF Conference on Computer Vision and Pattern
  Recognition (CVPR)}, pages 11082--11092, Los Alamitos, CA, USA, 2023. IEEE
  Computer Society.

\bibitem[Xu et~al.(2024)Xu, Bin, Wei, Yang, Wang, and
  Shen]{10568424_align_retrieve}
Yahui Xu, Yi Bin, Jiwei Wei, Yang Yang, Guoqing Wang, and Heng~Tao Shen.
\newblock Align and retrieve: Composition and decomposition learning in image
  retrieval with text feedback.
\newblock \emph{IEEE Transactions on Multimedia}, 26:\penalty0 9936--9948,
  2024.

\bibitem[Yang et~al.(2024)Yang, Xue, Qian, Dong, and Xu]{yang2024ldre}
Zhenyu Yang, Dizhan Xue, Shengsheng Qian, Weiming Dong, and Changsheng Xu.
\newblock Ldre: Llm-based divergent reasoning and ensemble for zero-shot
  composed image retrieval.
\newblock In \emph{Proceedings of the 47th International ACM SIGIR Conference
  on Research and Development in Information Retrieval}, pages 80--90, 2024.

\bibitem[Yao et~al.(2021)Yao, Huang, Hou, Lu, Niu, Xu, Liang, Li, Jiang, and
  Xu]{DBLP:journals/corr/abs-2111-07783_filip}
Lewei Yao, Runhui Huang, Lu Hou, Guansong Lu, Minzhe Niu, Hang Xu, Xiaodan
  Liang, Zhenguo Li, Xin Jiang, and Chunjing Xu.
\newblock {FILIP:} fine-grained interactive language-image pre-training.
\newblock \emph{CoRR}, abs/2111.07783, 2021.

\bibitem[Zeng et~al.(2023)Zeng, Wang, Yang, and Satoh]{zeng2023geo}
Zelong Zeng, Zheng Wang, Fan Yang, and Shin’ichi Satoh.
\newblock Geo-localization via ground-to-satellite cross-view image retrieval.
\newblock \emph{IEEE Transactions on Multimedia}, 25:\penalty0 2176--2188,
  2023.

\bibitem[Zhang et~al.(2024{\natexlab{a}})Zhang, Luan, Hu, Lee, Qiao, Chen, Su,
  and Chang]{zhang2024magiclens}
K. Zhang, Y. Luan, H. Hu, K. Lee, S. Qiao, W. Chen, Y. Su, and M-W. Chang.
\newblock {M}agic{L}ens: Self-supervised image retrieval with open-ended
  instructions.
\newblock In \emph{PMLR}, 2024{\natexlab{a}}.

\bibitem[Zhang et~al.(2024{\natexlab{b}})Zhang, Zhang, Chen, Wang, Chen, Xie,
  Sun, Deng, Zhang, Yang, Yang, Liao, Wang, and
  Guo]{zhang2024irgengenerativemodelingimage}
Yidan Zhang, Ting Zhang, Dong Chen, Yujing Wang, Qi Chen, Xing Xie, Hao Sun,
  Weiwei Deng, Qi Zhang, Fan Yang, Mao Yang, Qingmin Liao, Jingdong Wang, and
  Baining Guo.
\newblock Irgen: Generative modeling for image retrieval, 2024{\natexlab{b}}.

\end{thebibliography}
}

\end{document}